\documentclass[letterpaper, 10 pt, conference]{ieeeconf}  

\IEEEoverridecommandlockouts                              

\usepackage{amsmath} 
\usepackage{mathtools}

\usepackage{amsthm}
\usepackage{amssymb}

\usepackage{graphicx}
\usepackage{silence}
\usepackage{subcaption}
\usepackage{epstopdf}
\DeclareGraphicsExtensions{.eps}
\usepackage{booktabs}

\usepackage{verbatim} 

\newcommand{\id}[1][3]{{I}}
\newcommand{\zero}[2]{{0}}
\newcommand{\zeros}[2]{{0}}

\newcommand{\figref}[1]{{Figure~\ref{#1}}}

\newcommand{\multeqi}[2]{\begin{IEEEeqnarraybox}[][#2]{#1}}
\newcommand{\multeqf}{\end{IEEEeqnarraybox}}

\newcommand{\systemi}[1][rCL]{\left\lbrace\begin{IEEEeqnarraybox}[][c]{#1}}
\newcommand{\systemf}{\end{IEEEeqnarraybox}\right.}

\newcommand{\eqni}[1][rCL]{\begin{IEEEeqnarray}{#1}}
\newcommand{\eqnf}{\end{IEEEeqnarray}}

\newcommand{\nneqni}[1][rCL]{\begin{IEEEeqnarray*}{#1}}
\newcommand{\nneqnf}{\end{IEEEeqnarray*}}

\newcommand{\pmatrixi}{\begin{pmatrix}}
\newcommand{\pmatrixf}{\end{pmatrix}}

\newcommand{\bmatrixi}{\begin{bmatrix}}
\newcommand{\bmatrixf}{\end{bmatrix}}

\newcommand{\smatrixi}{\left[\begin{smallmatrix}}
\newcommand{\smatrixf}{\end{smallmatrix}\right]}

\newcommand{\enumi}{\begin{enumerate}}
\newcommand{\enumf}{\end{enumerate}}

\newcommand{\enumri}{\begin{enumerate}\renewcommand{\theenumi}{\textit{\roman{enumi}}}}
\newcommand{\enumrf}{\end{enumerate}}

\newcommand{\mytheorem}[2]{%
\newtheorem{t#2}{#1}%
\newenvironment{#2}{\begin{t#2}}{\end{t#2}}}

\theoremstyle{plain}

\mytheorem{Theorem}{theorem}
\mytheorem{Lemma}{lemma}
\mytheorem{Proposition}{proposition}
\mytheorem{Corollary}{corollary}
\mytheorem{Assumption}{assumption}
\mytheorem{Definition}{definition}
\mytheorem{Property}{property}

\title{\LARGE \bf
An Optimization Based Control Framework for Balancing and Walking: Implementation on the iCub Robot*
}

\author{Marie Charbonneau, Gabriele Nava, Francesco Nori, and Daniele Pucci$^{1}$
\thanks{*This work has received funding from the European Union’s Horizon 2020 research and innovation programme under the Marie Sklodowska-Curie grant agreement No 642667 (SECURE)}
\thanks{$^{1}$ Marie Charbonneau, Francesco Nori and Daniele Pucci are with the iCub Facility Department, Istituto Italiano di Tecnologia, Via Morego 30, 16163, Genova, Italy 
        {\tt\small (email: name.surname@iit.it)}
        }
}

\begin{document}

\maketitle
\thispagestyle{empty}
\pagestyle{empty}

\begin{abstract}

A whole-body torque control framework adapted for balancing and walking tasks is presented in this paper. In the proposed approach, centroidal momentum terms are excluded in favor of a hierarchy of high-priority position and orientation tasks and a low-priority postural task. More specifically, the controller stabilizes the position of the center of mass, the orientation of the pelvis frame, as well as the position and orientation of the feet frames. The low-priority postural task provides reference positions for each joint of the robot. Joint torques and contact forces to stabilize tasks are obtained through quadratic programming optimization. Besides the exclusion of centroidal momentum terms, part of the novelty of the approach lies in the definition of control laws in SE(3) which do not require the use of Euler parameterization. Validation of the framework was achieved in a scenario where the robot kept balance while walking in place. Experiments have been conducted with the iCub robot, in simulation and in real-world experiments.

\end{abstract}

\section{INTRODUCTION} \label{sec:introduction}
  
Robotic applications which are currently envisioned require not only that robots function autonomously in an environment adapted specifically to human capabilities, but also that robots coexist safely with humans. Humanoid robots can take advantage of the human form itself, in order to perform locomotion and physical interaction tasks with people or objects \cite{Hyon2007}. However, those physical interactions would influence stability and balance of the robot, and the development of robust compliant balancing controllers is therefore required. In this respect, torque-based controllers have shown to achieve motion control, while offering compliance and allowing control of physical interactions with the environment \cite{Saab2013, Ott2011}.

Various types of torque-based balancing controllers have been investigated specifically for biped robots. Fixed-based controllers considering a robot as a manipulator attached to the ground, either passivity-based \cite{Hyon2007}, bio-inspired \cite{Heremans2016} or momentum-based \cite{Ott2011}, have shown to ensure stable behavior. However, in order to achieve higher mobility, modeling a robot as a free-floating system, in which no link is fixed with respect to an inertial frame, would be relevant.

A framework enabling sequences of dynamic tasks for a free-floating torque-controlled humanoid robot has been developed in \cite{Salini2011}. Instead of solving a strict hierarchy of tasks, a single optimization problem was defined using a weighted combination of all tasks. Although this method allows for more flexibility in view of achieving all tasks, defining the value of the weights for each task may be a complex problem. As for control, an impedance controller was used to induce a desired behavior with respect to contacts with the environment, while an approximation of the zero moment point was used to perform balancing and walking tasks. Ultimately, important discontinuities were introduced when switching between constrained and unconstrained states, for which coping methods had to be proposed.

Instead, stability of a free-floating robot is typically ensured with momentum-based control strategies~\cite{Stephens2010, Herzog2014, Nava2016, Pucci2016Video}. These methods enable the stabilization of desired center of mass and contact wrenches through torque control. They have been used within task-based control strategies, aiming at the achievement of several control objectives organized in a hierarchical structure. The control objectives are generally formulated into optimization problems, to be solved recursively for each task and requiring the projection of low-priority task Jacobians in the null-space of higher-priority task Jacobians. Moreover, since task-based strategies allow the definition of control objectives with different priorities, specific motions could be achieved through the definition of lower-priority postural tasks. In \cite{lee2012} for example, desired center of mass and feet trajectories were taken into account for walking, allowing for specific motion of the swing foot. 

The generation of a centroidal momentum often relies on applying a torque about the center of mass, for instance by swinging the arms or bending quickly at the hips \cite{Stephens2010}. Although such behavior may enable a robot to maintain balance, the quick motions may not be ideal when interacting with humans. Moreover, joint limits of humanoid robots impose restrictions on movement. As a result, a desired angular momentum can typically be generated for short periods only, which may leave something to be desired, for example when subjected to continuous pushes. 

The whole-body controller presented in this paper was developed for balancing and it is formulated to be readily available for eventual dynamic locomotion tasks. 

The proposed method takes advantage of the task-based control strategy for balancing. However, instead of relying on a momentum-based strategy, the controller has for objective to stabilize the position and orientation of specific robot frames (center of mass position, pelvis orientation, feet position and orientation), while tracking joint positions. 

As such, this method opens possibilities in the choice of center of mass trajectory, as well as in the control of the orientations of body parts relevant to balancing and walking. In this way, the controller may explicitly act on the tendency of humans to draw on the control of pelvis and ankle orientation for balancing \cite{Horak1986, Stephens2007}. 

The controller relies on quadratic programming optimization in order to compute joint torques. As an additional contribution of the proposed method, the cost function is defined such that a single optimization problem is solved, for all tasks. Therefore, with this formulation, there is no need to solve lower-priority tasks projected into the null-space of higher priority tasks through a series of optimization problems.

It was implemented on the humanoid robot iCub \cite{Metta2008}, for a task consisting of repeatedly switching between double and single support: walking in place.

The paper is organized as follows. The next section introduces a formalism for modelling the system, as well as a generic formulation of task-based optimization problems. Section \ref{sec:background} also introduces control laws for position and orientation control. The control framework and its formulation are presented in section~\ref{sec:control_framework}. Experimental validation of the approach is discussed in \ref{sec:experimental_results}, followed by a conclusion opening on future work.

\section{BACKGROUND} \label{sec:background}

\subsection{Notation}

The following notation is used throughout the paper.
\begin{itemize}
	\item The set of real numbers is denoted by $\mathbb{R}$.
	\item $1_n \in \mathbb{R}^{n \times n}$ denotes the identity matrix of dimension $n$.
	\item $0_{n \times m} \in \mathbb{R}^{n \times m}$ is the zero matrix of dimension $n \times m$.
	\item The transpose operator is denoted by $(\cdot)^{\top}$.
	\item $S(\cdot)$ is the skew-symmetric operation associated with the cross product in $\mathbb{R}^3$.
	\item The vee operator denoted by $(\cdot)^{\vee}$ is the inverse of the $S(\cdot)$ operation, transforming a skew-symmetric matrix in $\mathbb{R}^{3 \times 3}$ into a vector in $\mathbb{R}^3$. 
	\item The Euclidean norm of a vector of coordinates $v \in \mathbb{R}^n$ is denoted by $\left|v\right|$.
	\item Given a time function $f(t) \in \mathbb{R}^n$, its first- and second-order time derivatives are denoted by $\dot{f}(t)$ and $\ddot{f}(t)$, respectively.
	\item $^{A}R_{B} \in SO(3)$ and $^{A}T_{B} \in SE(3)$ denote the rotation and transformation matrices which transform a vector expressed in the $B$ frame into a vector expressed in the $A$ frame.
\end{itemize}

\subsection{System Modelling}

It is assumed that the robot is composed of $n + 1$ rigid bodies, called links, connected by $n$ joints with one degree
of freedom each. The multibody system is considered free-floating, i.e. none of the links has an a priori
constant pose with respect to the inertial frame $\mathcal{I}$. The robot configuration space can then be characterized by the position and orientation of a frame attached to a link of the robot (called base frame $\mathcal{B}$) and the joint configurations. Thus, the configuration space is defined by $\mathbb{Q} = \mathbb{R}^3 \times SO(3) \times \mathbb{R}^n$. An element of $\mathbb{Q}$ is then a triplet $q = (^{\mathcal{I}}p_{B},~^{\mathcal{I}}R_{\mathcal{B}}, s)$, where $(^{\mathcal{I}}p_{\mathcal{B}},~^{\mathcal{I}}R_{\mathcal{B}})$ denotes the origin and orientation of the base frame expressed in the inertial frame, and $s$ denotes the joint angles. It is possible to define an operation associated with the set $\mathbb{Q}$ such that this set is a group. Given two elements $q$ and $\rho$ of the configuration space, the set $\mathbb{Q}$ is a group under the following operation: $q \cdot \rho = (p_q + p_{\rho}, R_qR_{\rho}, s_q + s_{\rho})$.

Furthermore, one easily shows that $\mathbb{Q}$ is a Lie group. Then, the velocity of the multibody system can be
characterized by the algebra $\mathbb{V}$ of $\mathbb{Q}$ defined by $\mathbb{V} = \mathbb{R}^3 \times \mathbb{R}^3 \times \mathbb{R}^n$. An element of $\mathbb{V}$ is then a triplet $\nu~=~(^{\mathcal{I}}\dot{p}_{\mathcal{B}},~^{\mathcal{I}}\omega_{\mathcal{B}}, \dot{s}) = (\text{v}_\mathcal{B}, \dot{s})$, where $^{\mathcal{I}}\omega_{\mathcal{B}}$ is the angular velocity of the base frame
expressed with respect to the inertial frame, i.e. $^{\mathcal{I}}\dot{R}_{\mathcal{B}} = S(^{\mathcal{I}}\omega_{\mathcal{B}}) ^{\mathcal{I}}R_{\mathcal{B}}$.

We also assume that the robot is interacting with the environment, exchanging $n_c$ distinct wrenches\footnote{As an abuse of notation, we define as \textit{wrench} a quantity that is not the dual of a \textit{twist}.}. The application
of the Euler-Poincar\'e formalism \cite[Chapter~13.5]{MechanicalSystemsBook} to the multibody system yields the following equations of motion:
\begin{IEEEeqnarray}{CCC}   \label{eq:system_dynamics}
    M(q) \dot{\nu} + h(q, \nu) =  
    \zeta \tau 
    + \displaystyle\sum_{k = 1}^{n_C} J_{c_k}^{\top} f_k
\end{IEEEeqnarray}
where $M \in \mathbb{R}^{n+6 \times n+6}$ is the mass matrix, $h \in \mathbb{R}^{n+6}$ is the bias vector of Coriolis and gravity terms, $\tau \in \mathbb{R}^n$ is a vector representing the actuation joint torques, $\zeta = (0_{n \times 6}, 1_n)^{\top}$ is a selector matrix, and $f_k \in \mathbb{R}^6$ denotes the $k$-th external wrench applied by the environment on the robot. We assume that the application point of the external wrench is associated with a frame $C_k$, attached to the link on which the wrench acts, and has its $z$ axis pointing in the direction of the normal of the contact plane. Then, the external wrench $f_k$ is expressed in a frame which has the same orientation as the inertial frame, and has its origin in $C_k$, i.e. the application point of the external wrench $f_k$.

The Jacobian $J_{c_k} = J_{c_k}(q)$ is the map between the robot velocity $\nu$ and the linear and angular velocities $^{\mathcal{I}}\text{v}_{C_k}~:=~(^{\mathcal{I}}\dot{p}_{C_k},~^{\mathcal{I}}\omega_{C_k})$ of the frame $C_k$, i.e. $^{\mathcal{I}}\text{v}_{C_k}~=~J_{c_k}(q)\nu$. The Jacobian has the following structure:
\begin{IEEEeqnarray}{CLL} \label{eq:contactJacobian}
    J_{C_k}(q) &=  
    \begin{bmatrix}
        J^b_{C_k} (q) & J^j_{C_k} (q)
    \end{bmatrix}
    & \in \mathbb{R}^{6 \times n + 6} \IEEEyessubnumber\label{eq:Jacobian_structure}\\
    J^b_{C_k}(q) &=
    \begin{bmatrix}
        1_{3} & -S(^{\mathcal{I}}p_{C_k} - ^{\mathcal{I}}p_{\mathcal{B}}) \IEEEyessubnumber\label{eq:Jacobian_base_structure}\\
        0_{3 \times 3} & 1_{3}
    \end{bmatrix}
    & \in \mathbb{R}^{6 \times 6} 
\end{IEEEeqnarray}

Lastly, it is assumed that holonomic constraints act on system~\eqref{eq:system_dynamics}. These constraints are of the form $c(q) = 0$ and may represent, for instance, a frame having a constant pose with respect to the inertial frame. In the case where this frame corresponds to the location at which a contact occurs on a link, we represent the holonomic constraint as:
\begin{equation}
    J_{C_k}(q)\dot{\nu} + \dot{J}_{C_k}(q,\nu)\nu = 0   \IEEEyessubnumber \label{eq:system_dynamics_contact_constraints}
\end{equation}

We then defined a control input $u$ to be composed of joint torques $\tau \in \mathbb{R}^n$ and a stacked vector of contact forces $F_c~\in~\mathbb{R}^{6i}$, where $i$ is the number of contacts, as follows:
\begin{equation}
    \label{eq:control_input}
	u = \begin{bmatrix} 
        \tau \\ F_c
    \end{bmatrix}
\end{equation}
In this case, for specifically handling single and double support on the left and right feet, we defined $F_C$ as
\begin{equation}
   F_C = \begin{bmatrix}
    F_L \\ F_R
   \end{bmatrix}
\end{equation}
where $F_L \in \mathbb{R}^6$ and $F_R \in \mathbb{R}^6$ are the contact wrenches at the left and right feet contacts. We also defined $J_C$ as
\begin{equation}
    J_C = \begin{bmatrix}
        \eta_L J_{C_L} \\
        \eta_R J_{C_R}
    \end{bmatrix}
\end{equation}
where $J_{C_L}$ and $J_{C_R}$ are the Jacobians associated to the left and right feet contacts, as defined in \eqref{eq:Jacobian_base_structure}; $\eta_L$ and $\eta_R$ are used as activation functions on the left and right feet contacts:
\begin{equation}
    \eta_x = \begin{cases}
        1 \text{ if foot $x$ is in contact with the ground} \\
        0 \text{ otherwise}
    \end{cases}
\end{equation}
From there, the system dynamics~\eqref{eq:system_dynamics} can be reformulated as
\begin{equation}
    \label{eq:system_dynamics_reformulated}
    M(q)\dot{\nu} + h = B u
\end{equation}
where
\begin{equation}
    \label{eq:B}
    B = 
    \left[
        \begin{array}{cc}  
            \zeta &
            J_C^T 
        \end{array}
    \right]
\end{equation}

\subsection{A classical optimization framework for balancing}

In the development of balancing controllers for humanoid robots, optimization techniques are commonly used in order to define control problem formulations taking constraints into account~\cite{Salini2011, Stephens2010, Herzog2014, Nava2016}. These formulations are often similar as exposed in \cite{Prete2016}, boiling down to a generic optimization problem of achieving desired behaviors by finding the input $u^*$ which minimizes a cost function given constraints:
\begin{subequations} 
    \label{eq:optimization_generic}
    \begin{align}
        u^* = \operatorname*{arg\,min}_u \text{  } & \left| A u - a \right|^2 \\ 
        \text{subject to  }
        & D_{eq} u + d_{eq} = 0\\
        & D u + d \leq 0 
    \end{align}
\end{subequations}
where the cost function represents the error on the task, which for torque controllers is generally defined to stabilize the centroidal momentum dynamics \cite{Stephens2010, Herzog2014, Nava2016, lee2012}. 

The equality constraint defined by $D_{eq}$ and $d_{eq}$ can be used to represent the system dynamics \eqref{eq:system_dynamics} and contact acceleration constraints \eqref{eq:system_dynamics_contact_constraints}. The inequality constraints defined by $D$ and $d$ can represent, among other things, torque limits, joint acceleration limits, or the force friction cones. Regarding the latter, the friction cone constraint shall be represented by an inequality of the form 
\begin{equation}
    C u \leq b
    \label{eq:friction_cone}
\end{equation}
with the matrix $C$ and vector $b$ chosen accordingly.

Furthermore, in the case of a multitask controller, the problem may be approached with two different strategies. The first one is based on a strict hierarchy between tasks, in which higher priority tasks would be fulfilled in priority, while the lower-priority tasks would be solved in the null space of the previous tasks \cite{Saab2013, Herzog2014, Nava2016}. The second strategy is instead based on weighted task priorities, where each task is associated with a weight defining its importance with respect to other tasks, and a solution is obtained from the combination of the weighted tasks \cite{Salini2011}.

Quadratic programming has been used in various contexts \cite{Heremans2016, Kuindersma2014, Escande2010}, offering a computationally efficient way to solve the inverse kinematics of complex systems in real-time. Therefore, the problem \eqref{eq:optimization_generic} is usually reformulated into quadratic programming (QP):
\begin{subequations}
    \label{eq:QP_generic}
    \begin{align}
        u^* = \operatorname*{arg\,min}_u \text{ }
        & \frac{1}{2} u^{\top} H u + u^{\top} g 
        \label{eq:QP_generic:argmin}\\
        \text{subject to }
        & \underbar{b} \leq A u \leq \bar{b} \label{eq:QP_generic:bounds}
    \end{align}
\end{subequations}
where the Hessian matrix $H$ is symmetric and positive definite and $g$ is the gradient vector. $A$ is the constraint matrix, with $\underbar{b}$ and $\bar{b}$ the associated lower and upper constraint vectors. 

\subsection{Proportional-derivative control in $SE(3)$}

The cost function and constraints associated to an optimization problem could be defined in terms of control laws. Given the configuration $(p, R) := (^{\mathcal{I}}p_{B},~^{\mathcal{I}}R_{B})$ of a frame $B$, this section proposes policies for stabilizing a desired position and orientation $(p_d, R_d)$ of a frame. The position and orientation problems are treated separately, in order to define desired linear and angular accelerations $(\ddot{p}^*, \dot{\omega}^*)~:= ~(^{\mathcal{I}}\ddot{p}_{B}^*,~^{\mathcal{I}}\dot{\omega}_{B}^*)$.

The desired linear acceleration is computed using the following proportional-derivative feedback control policy:
\begin{equation} \label{eq:linearPD}
    \ddot{p}^* =\ddot{p}_d - K_{P_l} (p - p_d) - K_{D_l} (\dot{p} - \dot{p}_d)
\end{equation}
where $K_{P_l} > 0$ and $K_{D_l} > 0$ are the linear proportional and derivative gains.

The problem of stabilizing a desired orientation $R$, on the other hand, may not be straightforward. For instance, the topology of $SO(3)$ forbids the design of smooth controllers that globally asymptotically stabilize a reference orientation \cite{Bhat2000}. Therefore, quasi-global asymptotic stability is commonly guaranteed by orientation controllers. This can be achieved through the following lemma, obtained from \cite[section 5.11.6, p.173]{Olfati2001}:
\begin{lemma}
    Let $\text{skew}(A) := \frac{1}{2}(A-A^{\top})$ for any matrix $A~\in~\mathbb{R}^3$. Consider the following orientation dynamics:
    \begin{equation}
        \begin{cases}
           \dot{R} = S(\omega)R \\
            \dot{\omega} = \dot{\omega}^{*}
        \end{cases}
    \end{equation}
where $\dot{\omega}^{*} \in \mathbb{R}^3$ is considered as control input. Assume that the control objective is the asymptotic stabilization of a desired attitude $(R_d(t),~\omega_d(t)) \in SO(3) \times \mathbb{R}^3$. Then,
\begin{subequations} \label{eq:rotationalPD}
\begin{align}
    ^{B}{\omega}_d ={}& R_d^{\top}\omega_d
    \label{eq:omega_B}
    \\
    \begin{split}
       ^{B}{\dot{\omega}}^{*} ={}& - K_{P_{\omega}} K_{D_{\omega}} \textnormal{skew}(R_d^{\top}R)^{\vee}\\
        & - K_{D_{\omega}} (^{B}{\omega}~-~^{B}{\omega}_d)\\
        & - K_{P_{\omega}} \textnormal{skew}(R_d^{\top}R ^{B}{\omega}^{\wedge}~-~^{B}{\omega}_{d}^{\wedge} R_d^{\top} R)^{\vee}
    \end{split}
    \label{eq:PD_rotation}
    \\
    \dot{\omega}^{*} ={}& R~^{B}{\dot{\omega}}^{*} + \dot{\omega}_{d}
    \label{eq:omega_star}
\end{align}
\end{subequations}
    renders the equilibrium point $(R, \omega) = (R_d, \omega_d)$ quasi-globally stable. $K_{P_{\omega}} > 0$ and $K_{D_{\omega}} > 0$ are angular proportional and derivative gains. 
\end{lemma}

To sum up, control of a frame in $SE(3)$ can be attempted using the control laws \eqref{eq:linearPD} and \eqref{eq:rotationalPD} yielding the desired frame acceleration.
\begin{equation}
    \dot{\text{v}}^* =
    \begin{bmatrix}
        \ddot{p}^* \\
        \dot{\omega}^*
    \end{bmatrix}
    \label{eq:PD_SE(3)}
\end{equation}

\section{CONTROL FRAMEWORK} \label{sec:control_framework}

This section exposes the proposed control framework, by defining the control objective for a whole-body balancing controller and how it was implemented as an optimization problem.

\subsection{Control objective}

\begin{figure}[!t]
    \centering
        \includegraphics[width=0.5\linewidth]{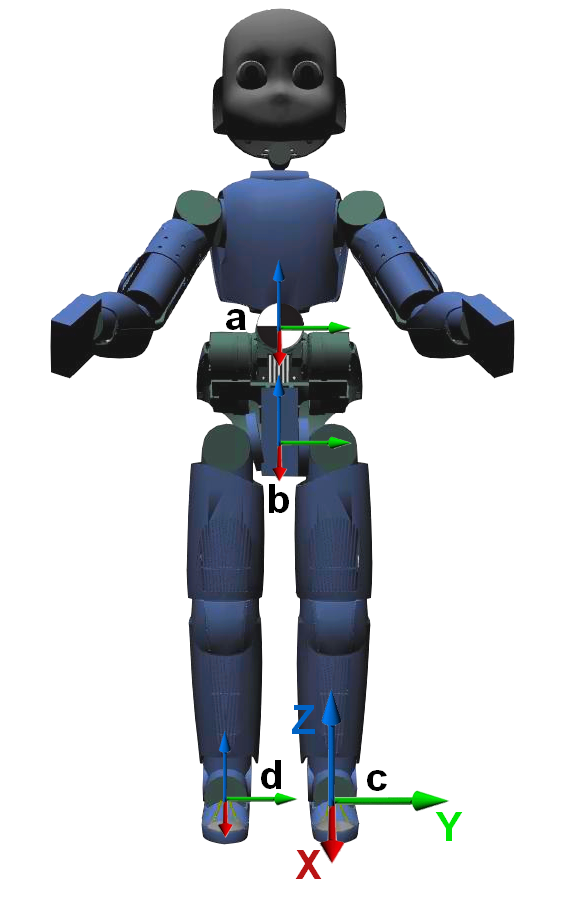}
    \caption{Frames considered in the control framework: \textbf{a} for the center of mass position, \textbf{b} for the root orientation, \textbf{c} for the left foot configuration and \textbf{d} for the right foot configuration; \textbf{c} also shows the x-y-z convention used in this paper.}    
    \label{fig:icub_frames}
\end{figure}

The developed controller has for objective to stabilize the center of mass position, root link orientation (to which the base frame $\mathcal{B}$ is attached, and which can be considered as the pelvis of the robot), as well as the left and right feet positions and orientations. The velocities associated to these four tasks are stacked together into $\Upsilon$:
\begin{equation}
    \label{eq:tasks}
	\Upsilon = \begin{bmatrix}
        \dot{p}_{G} \\ \omega_{\mathcal{B}} \\ \text{v}_L \\ \text{v}_R
    \end{bmatrix}
\end{equation}
where $\dot{p}_{G} \in \mathbb{R}^3$ is the linear velocity of the center of mass, $\omega_{\mathcal{B}} \in \mathbb{R}^3$ the angular velocity of the root frame, $\text{v}_L \in \mathbb{R}^6$ and $\text{v}_R \in \mathbb{R}^6$ are vectors of linear and angular velocities of the frames attached to the left and right feet. \figref{fig:icub_frames} illustrates the position of each of the concerned frames.
Letting $J_G$, $J_{\mathcal{B}}$, $J_L$, $J_R$ denote respectively the Jacobians of the center of mass position, root link orientation, left and right foot configurations, $J_{\Upsilon}$ can be defined as a stack of the Jacobians associated to each task:
\begin{equation}
    J_{\Upsilon} = 
    \begin{bmatrix}
        J_{G} \\
        J_{{\mathcal{B}}} \\
        J_{L} \\
        J_{R} \\
    \end{bmatrix}
\end{equation}

Furthermore, the task velocities $\Upsilon$ can be computed from $\nu$ using $\Upsilon = J_{\Upsilon} \nu$. By deriving this expression, the task acceleration is
\begin{equation}
    \label{eq:task_acceleration}
    \dot{\Upsilon} = \dot{J}_{\Upsilon} \nu + J_{\Upsilon} \dot{\nu}
\end{equation}

In view of~\eqref{eq:system_dynamics} and~\eqref{eq:task_acceleration}, the task accelerations $\dot{\Upsilon}$ can be formulated as a function of the control input $u$:
\begin{equation}
    \label{eq:task_acceleration_from_u}
    \dot{\Upsilon}(u) = \dot{J}_{\Upsilon} \nu + J_{\Upsilon} M^{-1} (B u - h)
\end{equation}

Finally, we also added the lower priority postural task of tracking joint configurations. Similarly as above, in view of~\eqref{eq:system_dynamics_reformulated}, one can obtain $\ddot{s}(u)$, a formulation of the joint accelerations $\ddot{s}$ as a function of the control input $u$:
\begin{equation}
    \label{eq:sdd_from_u}
    \ddot{s}(u) = 
    \zeta^{\top} 
    \left[ M^{-1} (Bu - h) \right]
\end{equation}

\subsection{Strict tasks formulation}

The application of strict task priorities in the present case would enforce the position and orientation tasks to be fulfilled in priority, while the lower-priority postural task is to be optimized depending on its feasibility. Therefore, the input $u^*$ would be obtained by minimizing the joint tracking error, while satisfying constraints enforcing the values associated to the joint and task accelerations, as well as the friction cone constraints. Applying the formulation of \eqref{eq:optimization_generic}, we defined:
\begin{subequations} 
    \label{eq:optimisation_strict_priorities}
    \begin{align}
        u^* = \operatorname*{arg\,min}_u \text{ }
        & \frac{1}{2} \left| \ddot{s}(u) - \ddot{s}^* \right| ^2 \label{eq:optimisation_strict_priorities:argmin}\\
        \text{subject to }
        & \ddot{s}^* = -K_{P_s} (s - s_d) - K_{D_s} (\dot{s} - \dot{s}_d) \label{eq:optimisation_strict_priorities:joint_constraints} \\ 
        & \dot{\Upsilon}^* = \dot{\Upsilon}(u) \label{eq:optimisation_strict_priorities:task_constraints}\\
        & C u \leq b \label{eq:optimisation_strict_priorities:friction_cone_constraint}
    \end{align}
\end{subequations}

The postural task~\eqref{eq:optimisation_strict_priorities:joint_constraints} is achieved through a computed-torque-like control strategy at the joint level, which stabilizes joint positions toward a desired posture $s_d$ by imposing joint torques $\tau^*$ computed from desired joint accelerations $\ddot{s}^*$ and the system dynamics~\eqref{eq:system_dynamics}:
\begin{equation}
    \label{eq:computed-torque}
    \tau^* = M \ddot{s}^* + h
\end{equation}
We then compute $\ddot{s}^*$ in~\eqref{eq:optimisation_strict_priorities:joint_constraints} from the error on joint positions and velocities, where $K_{P_s}$ and $K_{D_s}$ are proportional and derivative gains associated to the joints. 

The second constraint equation~\eqref{eq:optimisation_strict_priorities:task_constraints} was defined from~\eqref{eq:task_acceleration_from_u}, given desired task accelerations $\dot{\Upsilon}^*$ obtained from \eqref{eq:linearPD}, \eqref{eq:rotationalPD} or \eqref{eq:PD_SE(3)} depending on each task:
\begin{equation}
    \dot{\Upsilon}^* = \begin{bmatrix}
        \ddot{p}^*_{G} \\
        \dot{\omega}^*_{{\mathcal{B}}} \\
        \dot{\text{v}}^*_{L} \\
        \dot{\text{v}}^*_{R}
    \end{bmatrix}
    \label{eq:task_accelerations}
\end{equation}

The third constraint~\eqref{eq:optimisation_strict_priorities:friction_cone_constraint} was defined as in \eqref{eq:friction_cone} to keep contact forces within the associated friction cone.

Using this formulation, there is a possibility that the lower-priority postural task may not be achieved at all, resulting in a movement which is not satisfactory in terms of the global desired behavior.

\subsection{Weighted tasks formulation}

In the present context where tasks may have relative priorities, soft task priorities would allow to achieve a trade-off between weighted tasks. The problem can thus be reformulated with soft task priorities, by using a weighted sum to integrate the equality constraint~\eqref{eq:optimisation_strict_priorities:task_constraints} into the cost function:
\begin{subequations}
    \label{eq:optimisation_weighted_tasks}
    \begin{align}
        u^* = \operatorname*{arg\,min}_u \text{ }
        & \frac{w_{s}}{2} \left| \ddot{s}(u) - \ddot{s}^* \right| ^2
        + \frac{w_{\Upsilon}}{2} \left| \dot{\Upsilon}(u) - \dot{\Upsilon}^* \right|^2 
        \label{eq:optimisation_weighted_tasks:argmin}\\
        \text{subject to }
        & \ddot{s}^* = - K_{P_s} (s - s_d) - K_{D_s} (\dot{s} - \dot{s}_d) \label{eq:optimisation_weighted_tasks:joint_constraints}\\
        & \dot{\Upsilon}^* =  \begin{bmatrix}
        \ddot{p}^*_{G} \\
        \dot{\omega}^*_{\mathcal{B}} \\
        \dot{\text{v}}^*_{L} \\
        \dot{\text{v}}^*_{R}
    \end{bmatrix} \label{eq:optimisation_weighted_tasks:task_constraints}  \\ 
        & C u \leq b \label{eq:optimisation_weighted_tasks:friction_cone_constraint}
    \end{align}
\end{subequations}

In this formulation, $w_{s}$ and $w_{\Upsilon}$ are the weights associated to posture and task costs, respectively. Since task tracking is of high priority in the case of the balancing controller, $w_{\Upsilon}$ shall be attributed the highest value. 
 
\subsection{Quadratic programming formulation}

The optimization problems obtained in either~\eqref{eq:optimisation_strict_priorities} or~\eqref{eq:optimisation_weighted_tasks} can be transformed into a quadratic programming formulation of the form \eqref{eq:QP_generic}. The Hessian matrix $H$ and gradient vector $g$ can be obtained from~\eqref{eq:B},~\eqref{eq:task_acceleration_from_u},~\eqref{eq:sdd_from_u} and the previously formulated optimization~\eqref{eq:optimisation_strict_priorities} or~\eqref{eq:optimisation_weighted_tasks}. 

The constraint matrix A is obtained from C in the friction cone constraints~\eqref{eq:optimisation_strict_priorities:friction_cone_constraint} or~\eqref{eq:optimisation_weighted_tasks:friction_cone_constraint}, and from the task accelerations equality constraint~\eqref{eq:optimisation_strict_priorities:task_constraints} in the case of strict task priorities. Regarding lower and upper bounds $\underbar{b}$ and $\bar{b}$, the friction cone constraints are considered without a lower bound, while the constraint~\eqref{eq:optimisation_strict_priorities:task_constraints}, on the other hand, is considered bounded to a single value.

\section{SIMULATION AND EXPERIMENTAL RESULTS} \label{sec:experimental_results}

The proposed control law was tested in simulation and in experiments with the humanoid robot iCub. The conducted experiments consisted in balancing on two feet, then repeatedly switching between double and single support by lifting one foot and the other by 5~cm, as if walking in place. Moreover, the foot was kept in its lifted position for a duration of 5 seconds in simulation, 15 seconds on the robot.

Before exposing achieved results, a few notes are added on how the proposed control framework was implemented. 

\subsection{Implementation of the balancing controller}
\begin{figure}[!t]
    \centering
    \vspace{0.5em}
        \includegraphics[width=\linewidth]{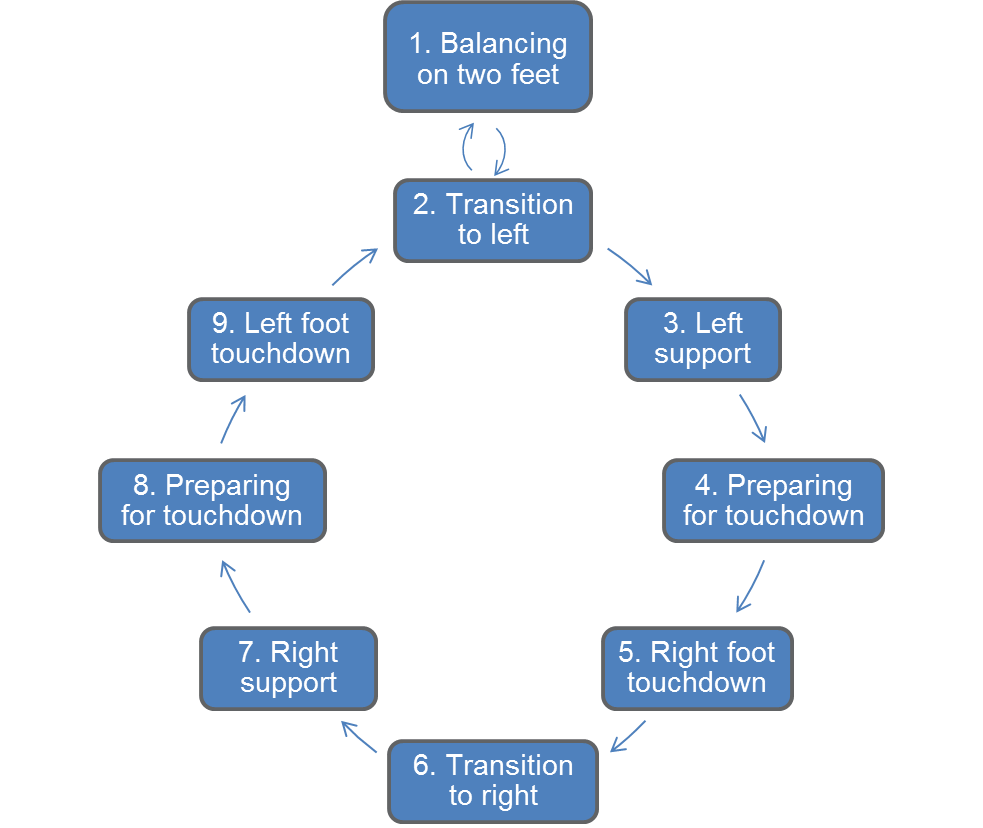}
    \caption{State machine used for generating walking motion.}    
    \label{fig:state_machine}
\end{figure}  

\begin{figure}[!t]
    \centering
    
    \begin{subfigure}[b]{0.48\textwidth}
        \includegraphics[width=0.19\linewidth]{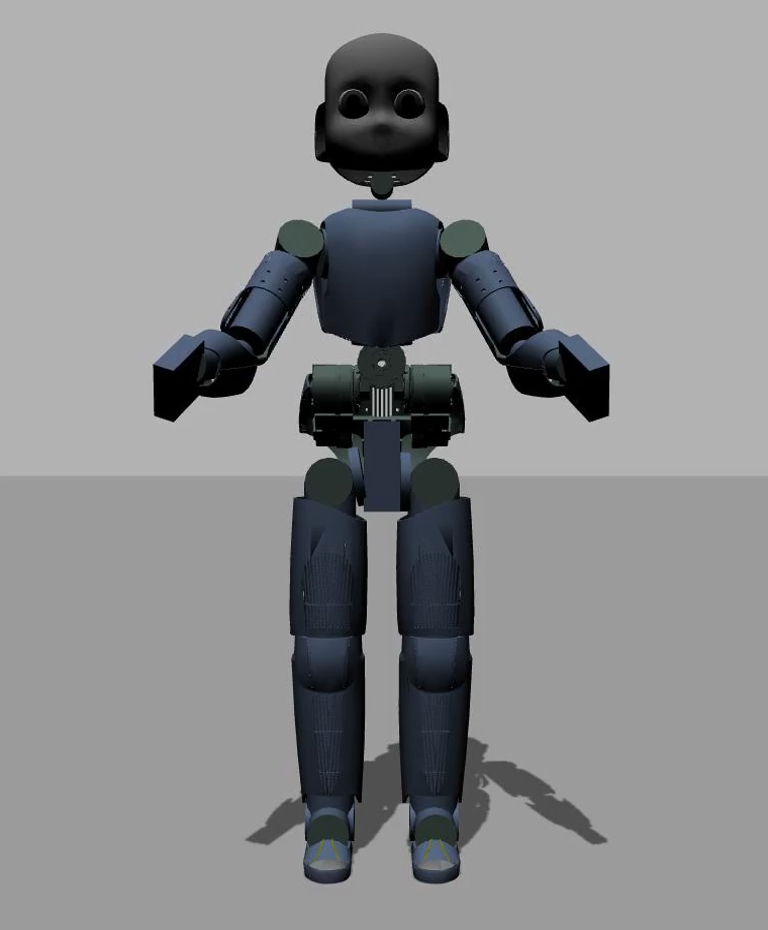} 
        \includegraphics[width=0.19\linewidth]{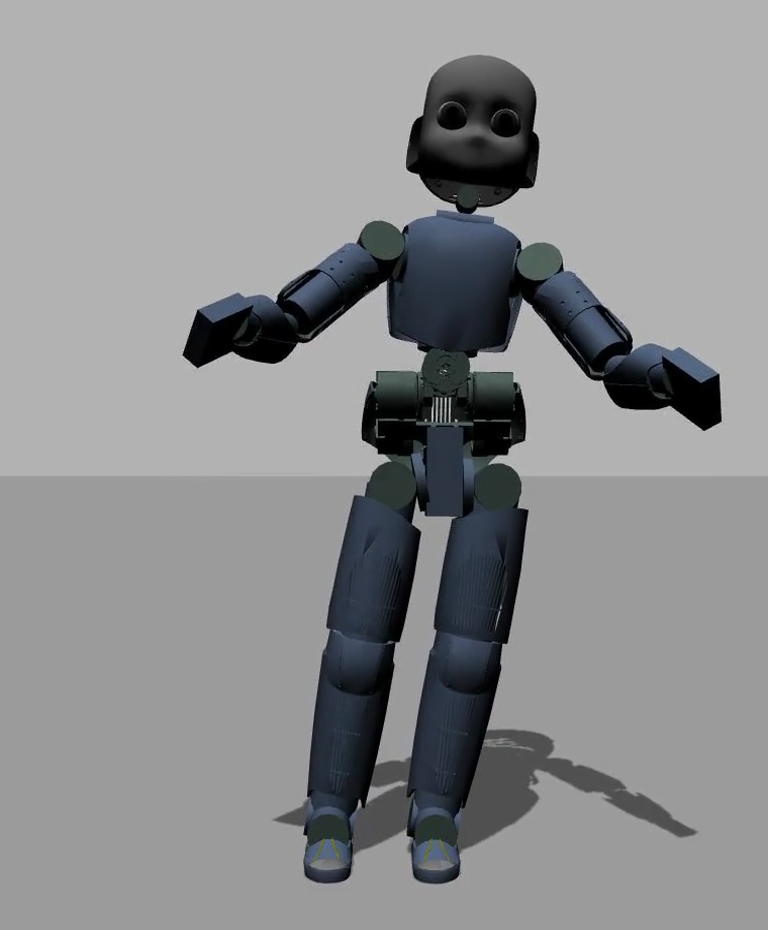} 
        \includegraphics[width=0.19\linewidth]{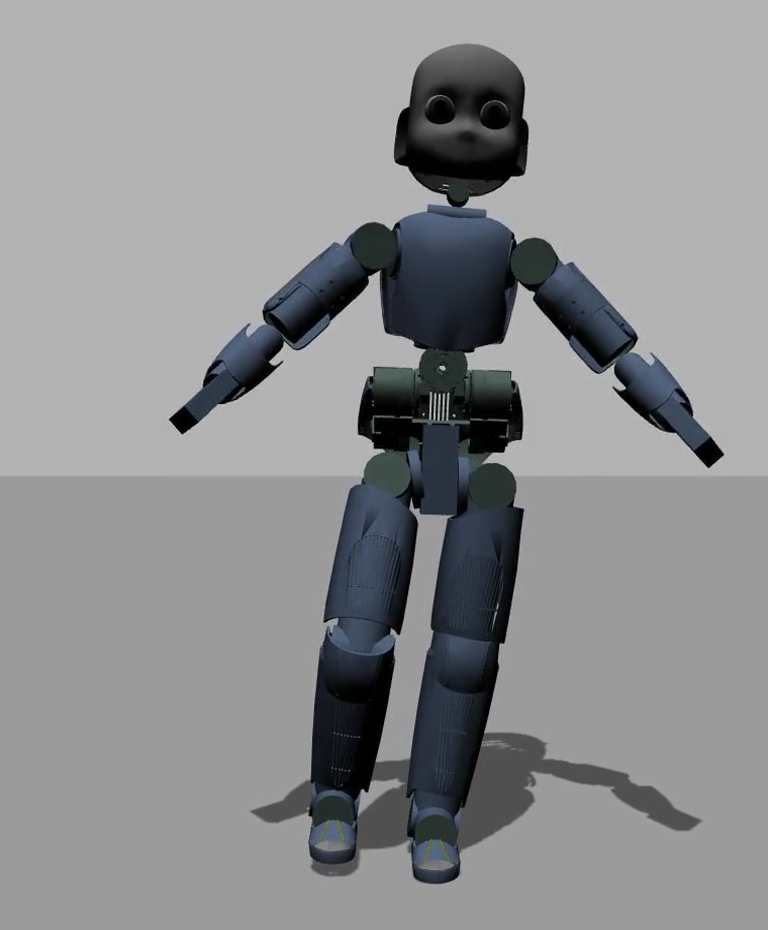} 
        \includegraphics[width=0.19\linewidth]{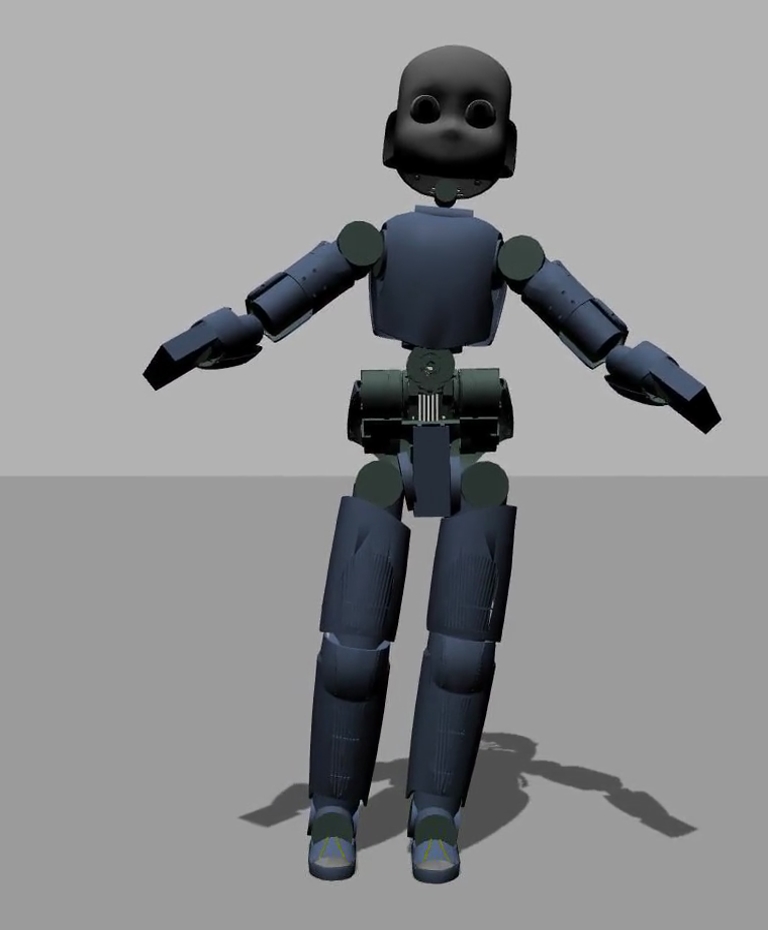} 
        \includegraphics[width=0.19\linewidth]{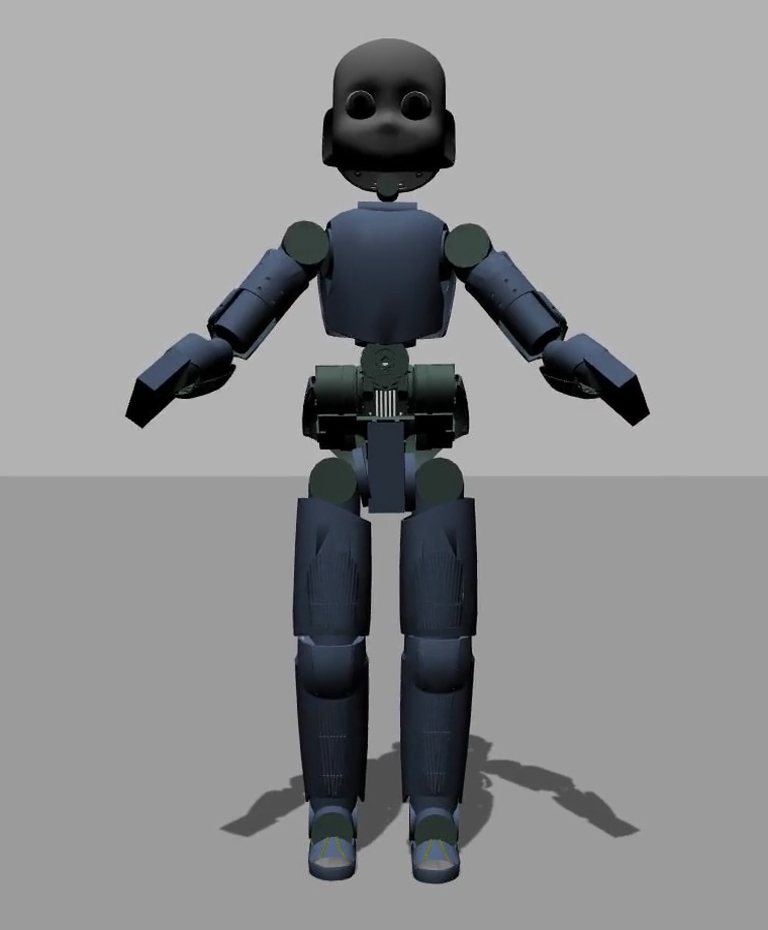} 
    \end{subfigure}\\
    \vspace{0.1cm}
    \begin{subfigure}[b]{0.48\textwidth}
        \includegraphics[width=0.19\linewidth]{images/simulation/vlcsnap-00011.jpg} 
        \includegraphics[width=0.19\linewidth]{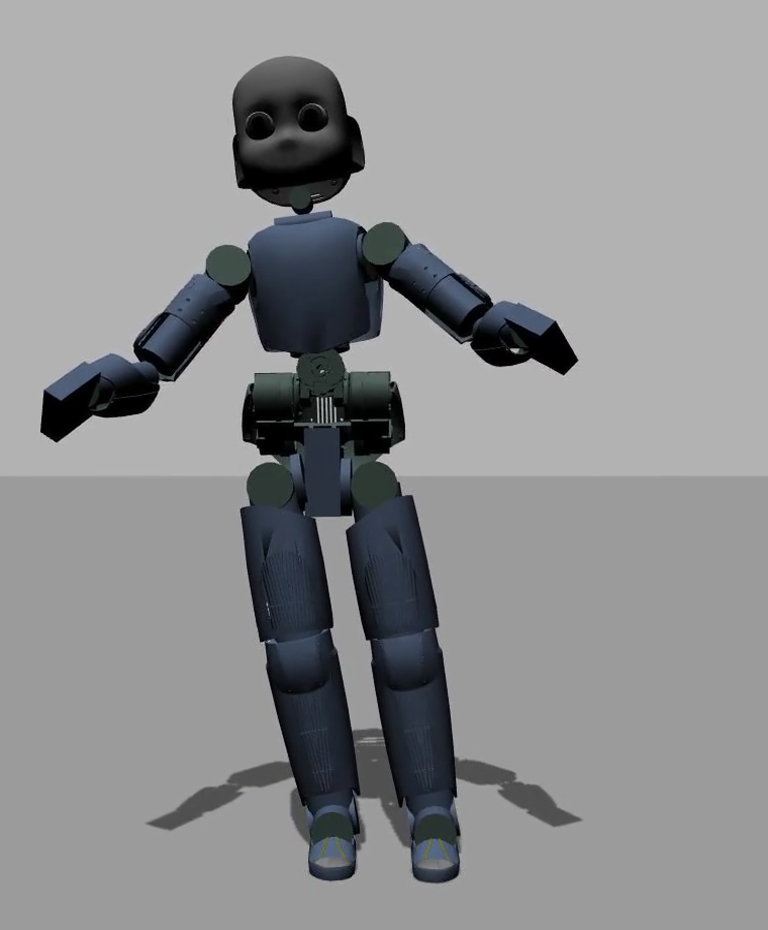} 
        \includegraphics[width=0.19\linewidth]{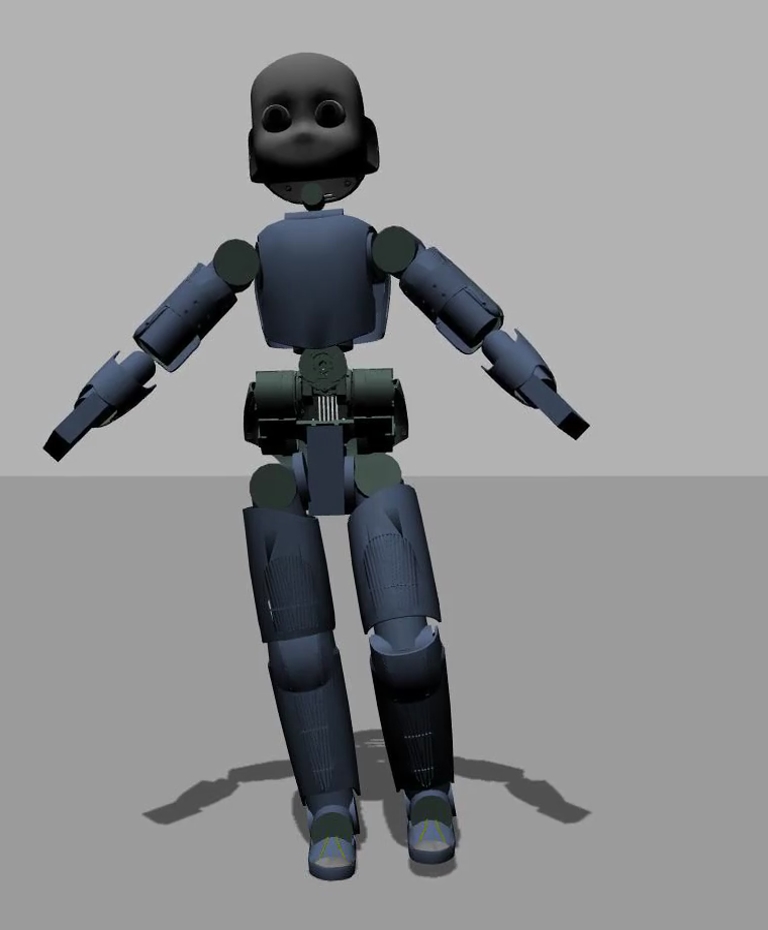} 
        \includegraphics[width=0.19\linewidth]{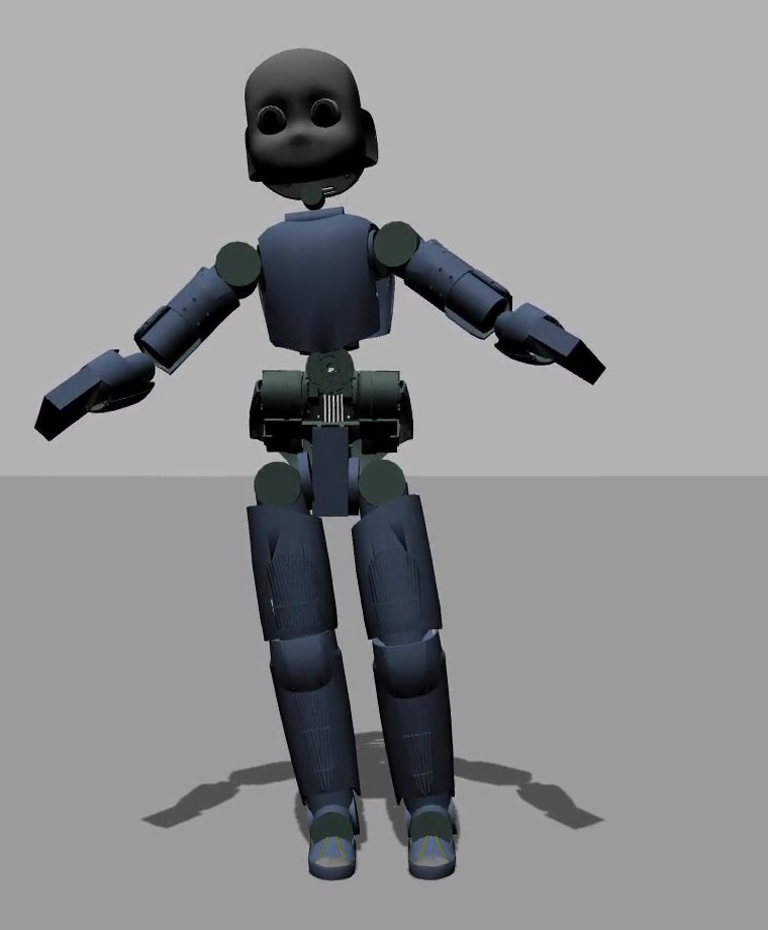} 
        \includegraphics[width=0.19\linewidth]{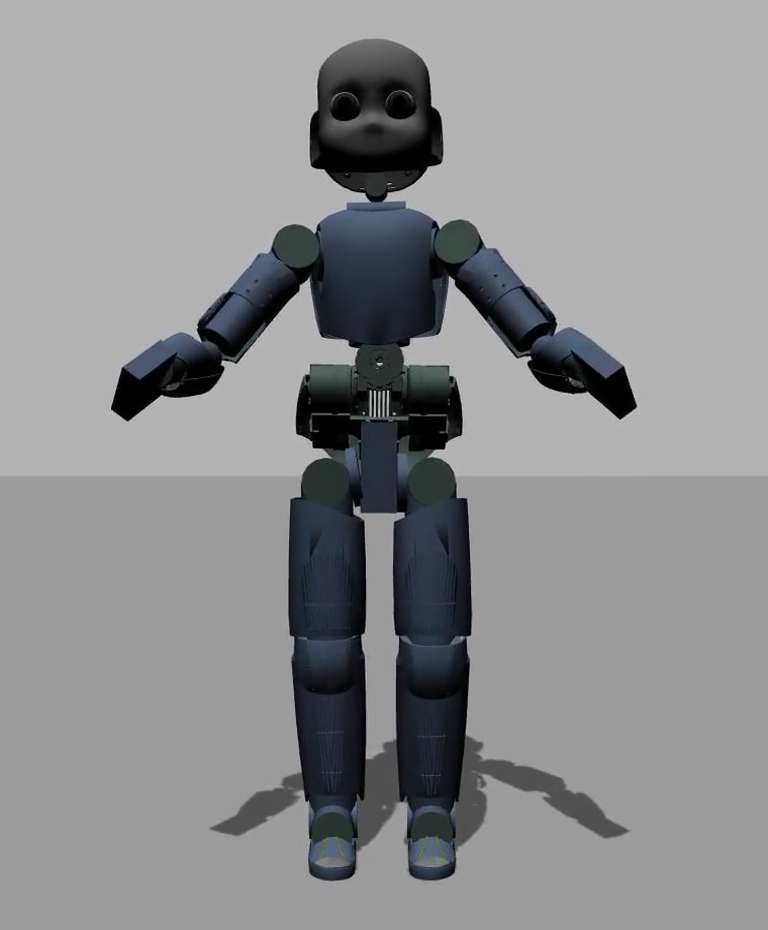} 
    \end{subfigure} 
    
    \caption{Walking in place motion achieved with the proposed controller: lifting the right foot, then the left foot.}
    \label{fig:snapshots}
\end{figure}  

The weighted tasks approach~\eqref{eq:optimisation_weighted_tasks} was retained for the implementation presented in this paper. Task weights were kept constant for the scope of this work, but adapting weights over time, depending on the sequence of actions of the robot, could eventually be developed in order to improve the global behavior of the robot \cite{Liu2015, Modugno2016}.

An additional constraint was added to \eqref{eq:optimisation_weighted_tasks}, to ensure a measure of continuity in the joint torques. It was defined by applying a maximum rate of change $\dot{\tau}_{max}$ to the torque value computed at the previous time step $\zeta_{u} u_{(t-1)}$:
\begin{equation} \label{eq:continuity_constraint}
    -\dot{\tau}_{max} \Delta_t + \zeta_{u} u_{(t-1)} \leq \zeta_{u} u \leq \dot{\tau}_{max} \Delta_t + \zeta_{u} u_{(t-1)}
\end{equation}
where $\Delta_t$ is the duration of a time step. The selector matrix $\zeta_{u} = (1_n, 0_{n \times 6})$ is used to specifically target the torques. 

Furthermore, a regularizing term was added in \eqref{eq:optimisation_weighted_tasks:argmin}, in order to encourage solutions with lower torque amplitudes. It is important to note here that we regularized only $\zeta_{u} u$, the joint torques portion of the control input. As explained in \cite{Romano2017}, minimizing contact wrenches would have the undesirable effect of enforcing an almost constant vertical force at the contacts, independently of the center of mass position.

A state machine as illustrated in~\figref{fig:state_machine} was used to output desired setpoints for each task and each joint, in function of the state of the robot. The state machine was also used for gain scheduling. Transition from one state to the next was smoothed with a minimum jerk trajectory~\cite{Pattacini2010}.

Concerning the movement of vertically lifting the foot by 5~cm, the iCub can achieve it by a simple rotation of the root link, while keeping the legs straight. This is a straightforward solution which requires a limited amount of torques, and as a result it would be promoted by the quadratic programming. In order to achieve a more human-like behavior, the following additional tasks were defined for the states where a foot was lifted (left or right support). The first one was to keep the root link orientation parallel to the stance foot, in order to discourage the solution mentioned above. The second was to keep the orientation of the foot parallel to the stance foot, in order to ensure that all points of the foot are lifted equally. Finally, a postural task was used, requiring to bend the knee of the lifted leg. 

The same could be discussed for foot touchdown: lowering the root link would limit the amount of torque used to bring the foot to the ground, compared to unbending the hip, knee and ankle. By keeping the root link orientation constant during the state where the foot was lowered (preparing for touchdown), one could enforce the straightening of the leg.

The implemented controller runs in real-time, generating joint torque commands every millisecond. It is using the open-source software package qpOASES~\cite{Ferreau2014}, for solving the quadratic programming. The full code, including the definition of all control parameters, is available online \cite{ImplementationCode}.

\subsection{Simulation results}

Experiments were conducted in simulation using the open-source robot simulator Gazebo \cite{Koenig2004}. The achieved motion is illustrated in~\figref{fig:snapshots}: the robot begins on double support, then transitions to single support on the left foot, before lifting the right foot and lowering it back to the ground and repeating the same process on the other side. It can be noted that the root link and feet remain horizontal, as required by the orientation tasks. The robot could repeat the process of lifting one foot after the other practically indefinitely, without loss of balance: it was validated in simulation for a minimum of 75 continuous cycles. However, the results presented here are limited to showing the first cycle.

\begin{figure}
    \centering
    
    \begin{subfigure}[b]{0.48\textwidth}
    \includegraphics[width=0.3\linewidth]{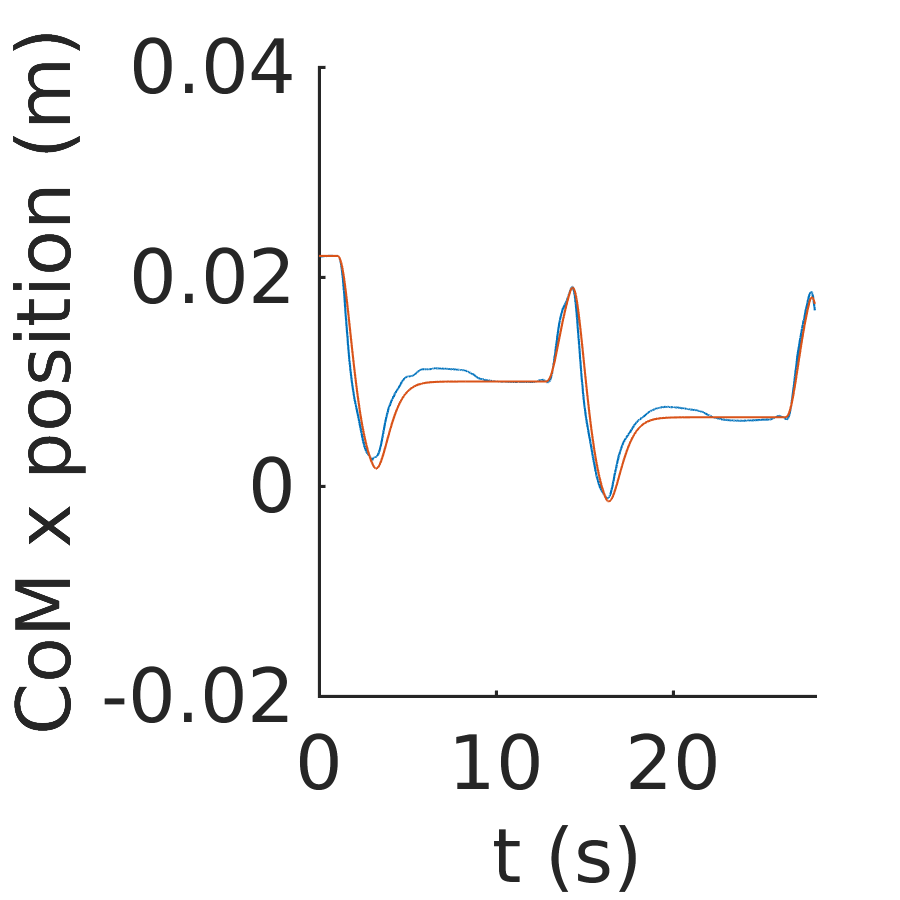}
    \includegraphics[width=0.3\linewidth]{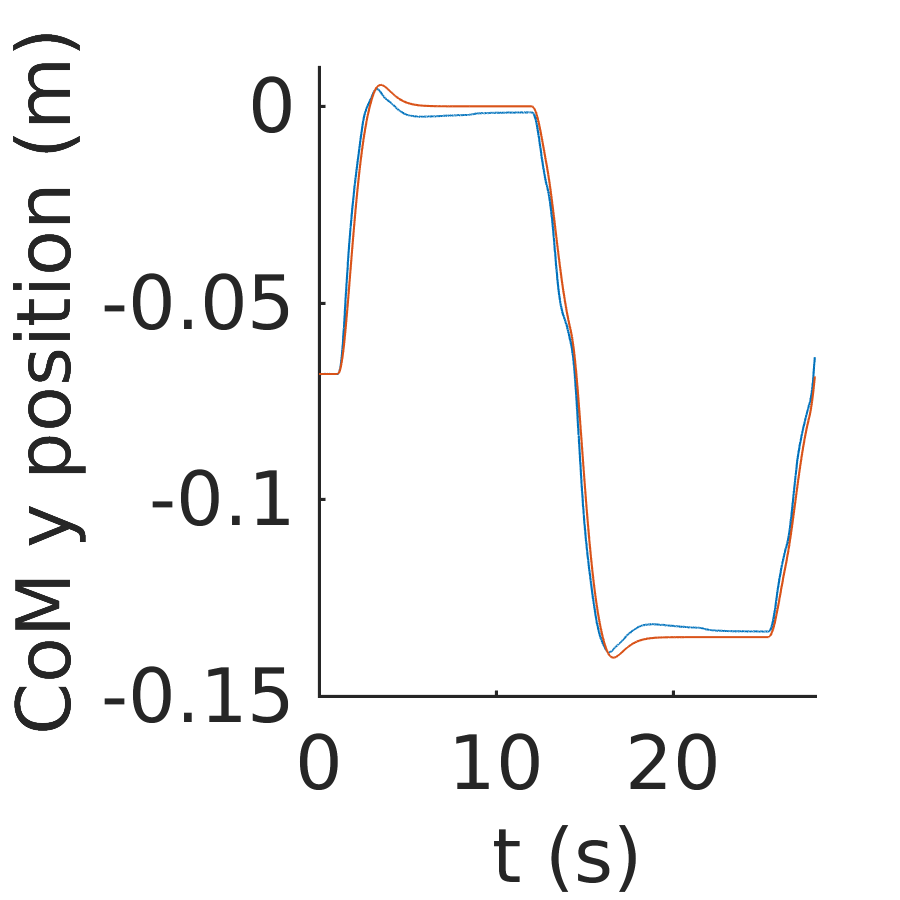}
    \includegraphics[width=0.3\linewidth]{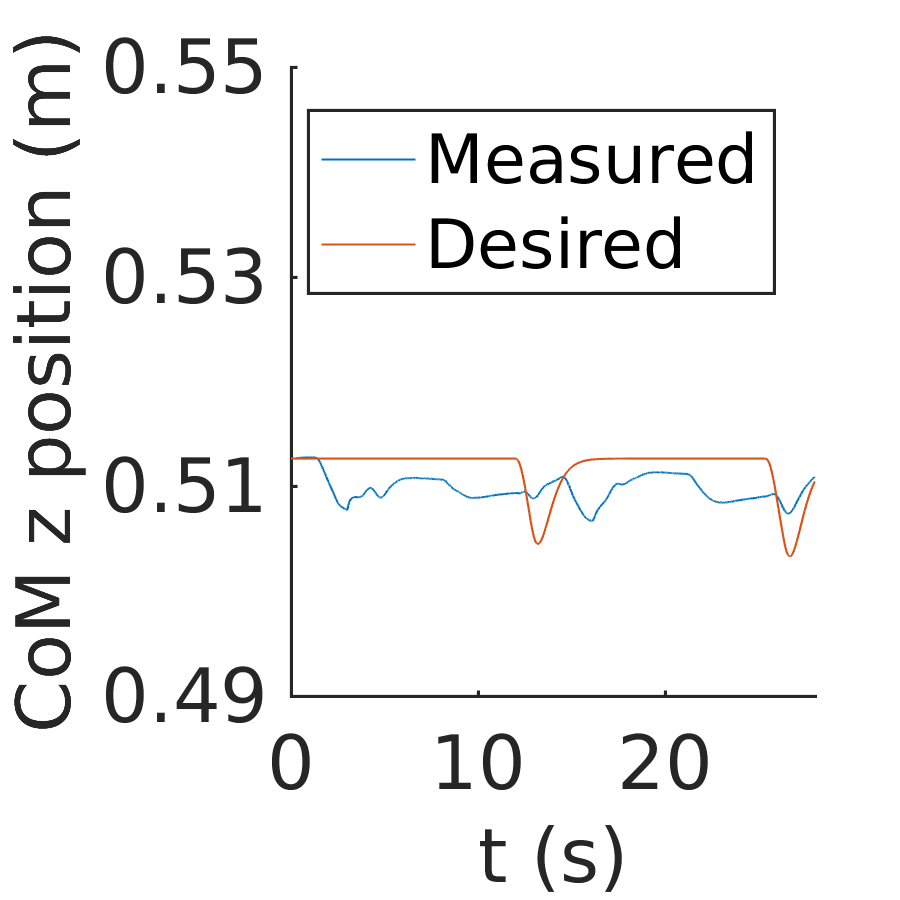}
    \caption{Center of mass (CoM) position}
    
    \includegraphics[width=0.3\linewidth]{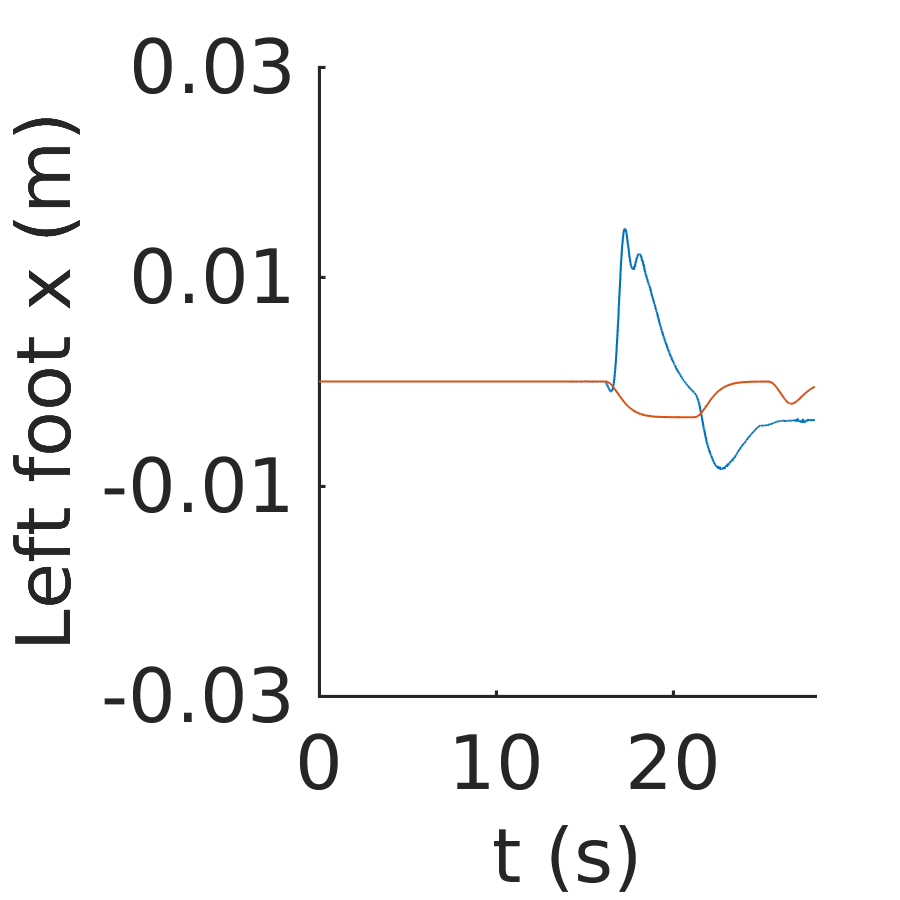}
    \includegraphics[width=0.3\linewidth]{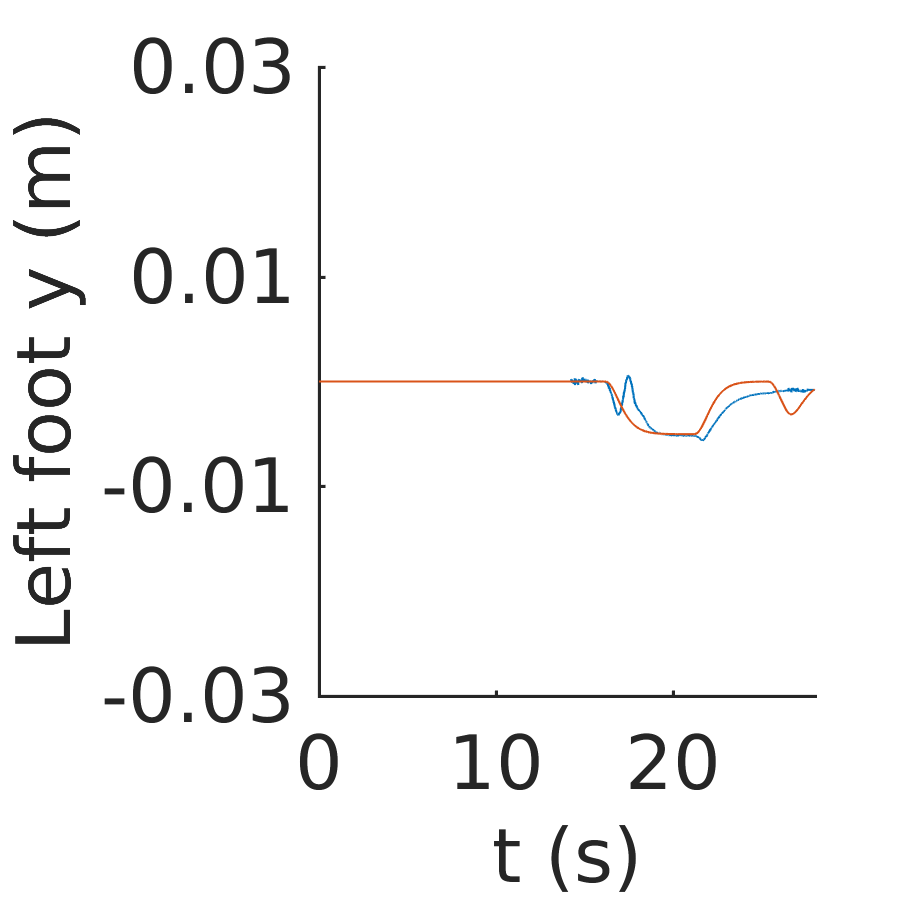}
    \includegraphics[width=0.3\linewidth]{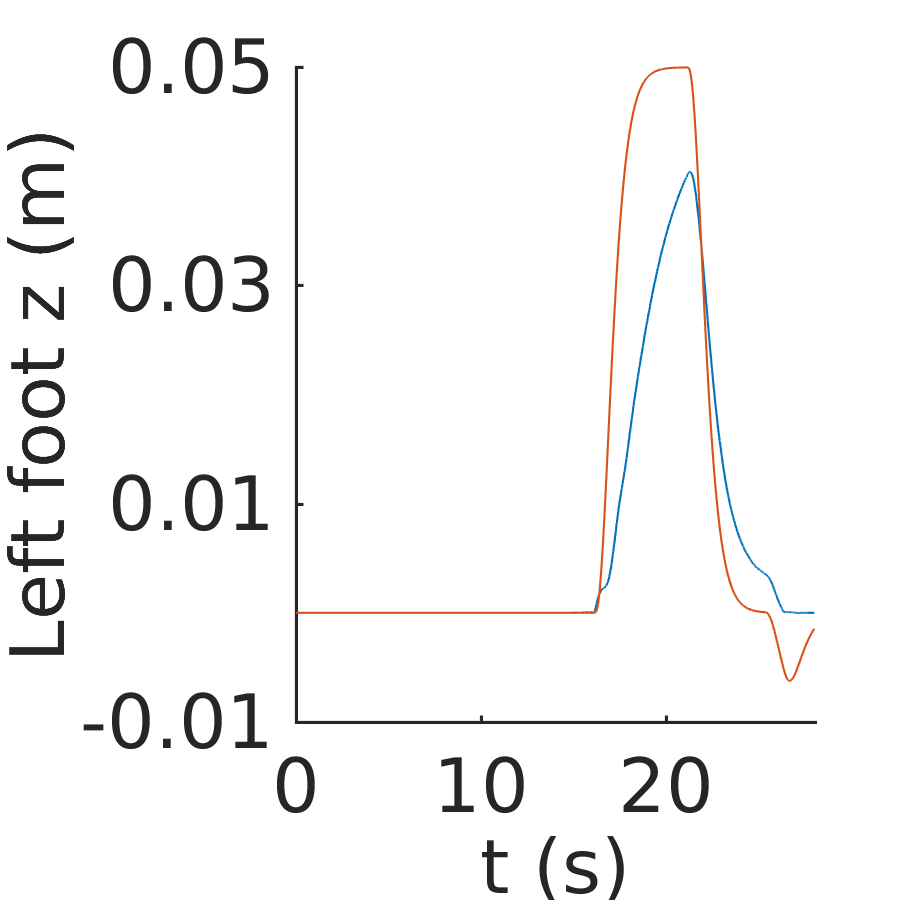}
    \caption{Left foot position}
    
    \includegraphics[width=0.3\linewidth]{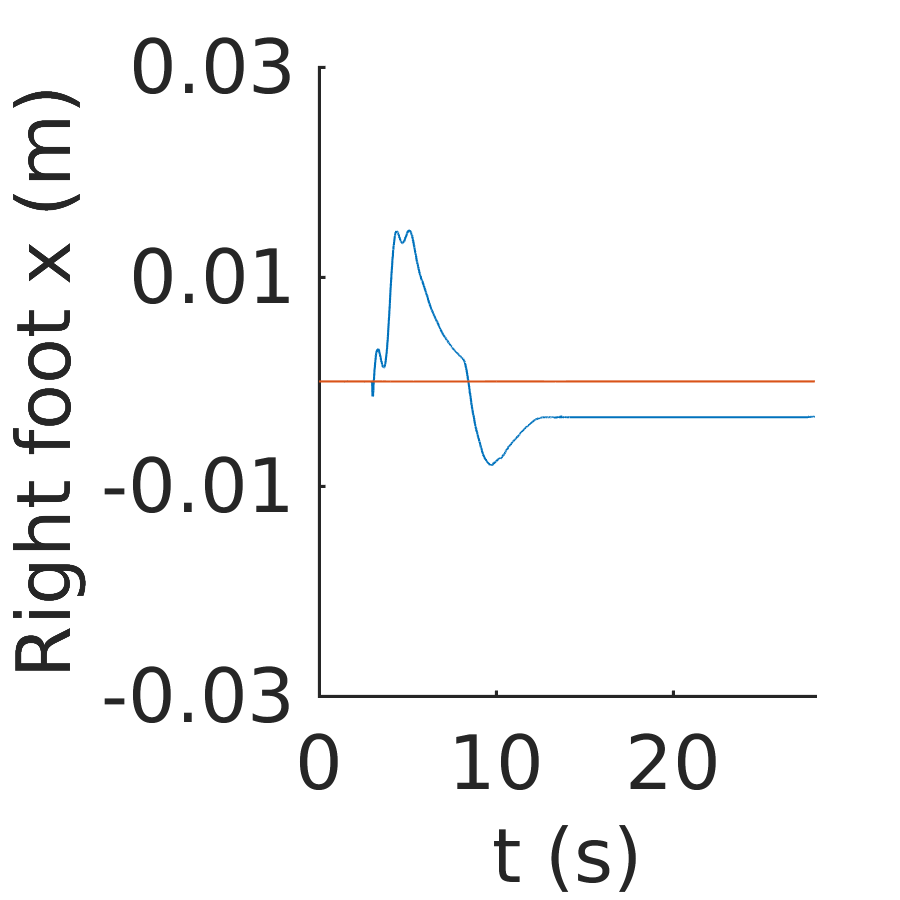}
    \includegraphics[width=0.3\linewidth]{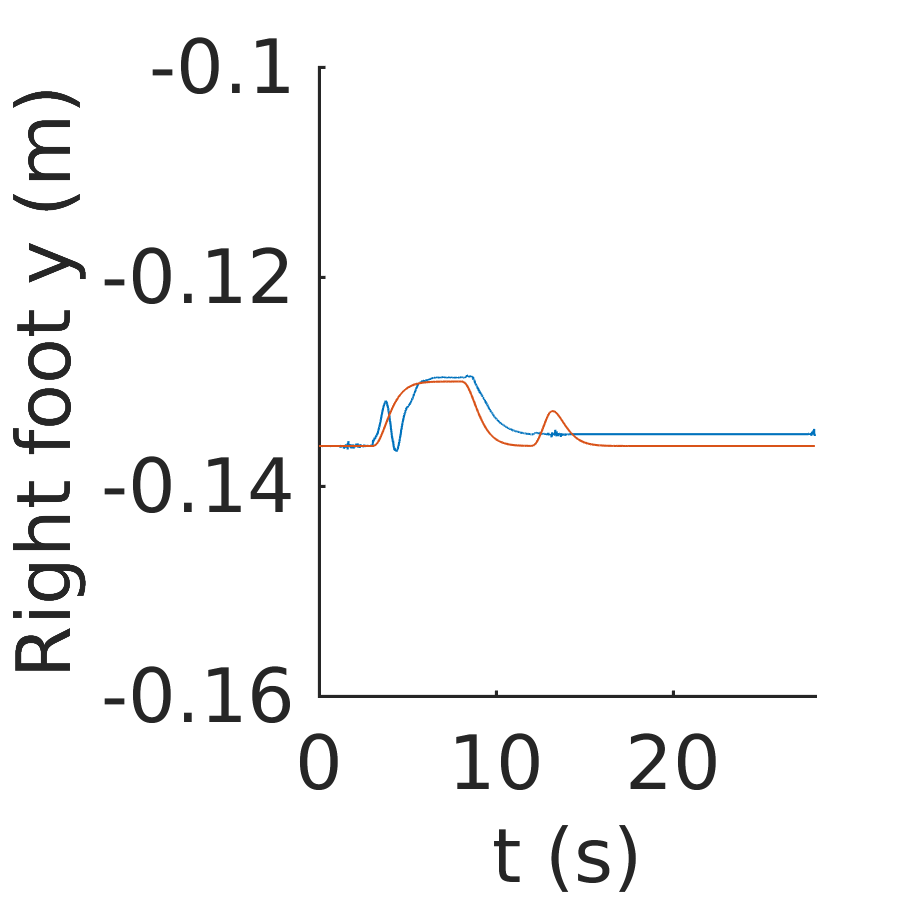}
    \includegraphics[width=0.3\linewidth]{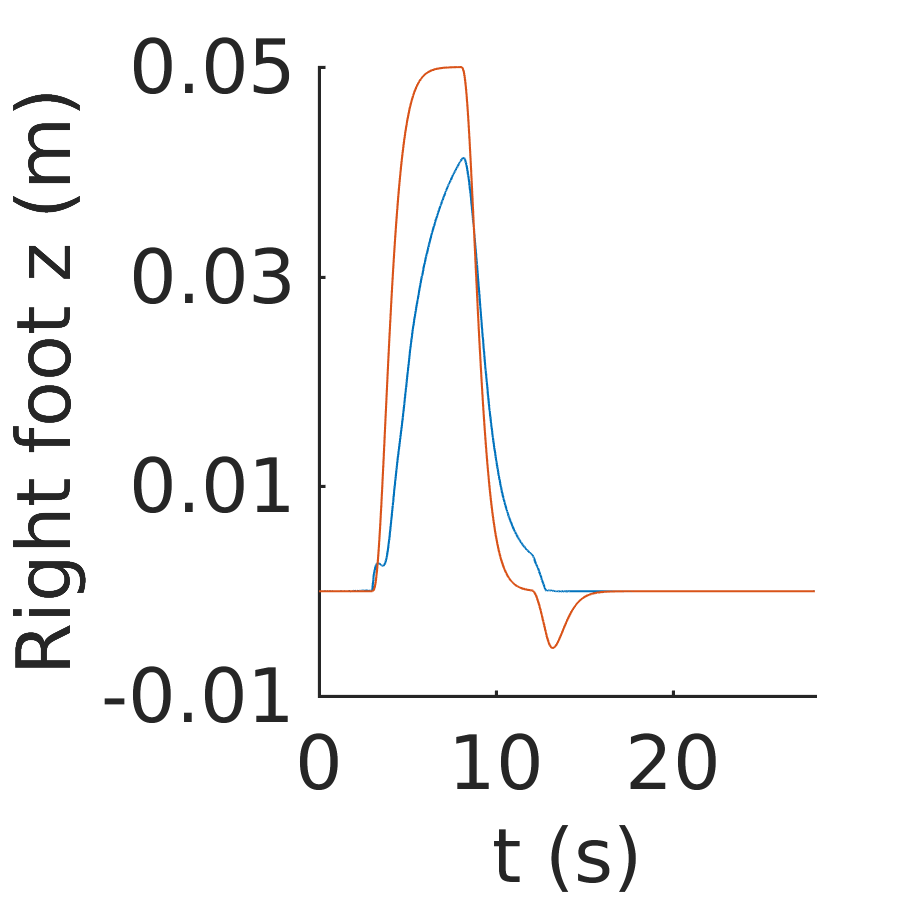}
    \caption{Right foot position}
    \end{subfigure}

    \caption{Evolution of position tasks for a sample of 1 stride, achieved in simulation. Position values are given with respect to a world frame of which the $x$, $y$ and $z$ axes correspond respectively to the sagittal, frontal and vertical axes. Achieved trajectories are shown in blue, while the desired ones are shown in red.}
    \label{fig:task_evolution}
\end{figure}

The trajectories obtained for the center of mass and feet are shown in~\figref{fig:task_evolution}. The error on the center of mass was generally kept below 0.01~m at all times. Although the feet positions were treated with a lower priority than the center of mass (through the use of lower proportional and derivative gains), the maximum tracking error, which was obtained on the vertical axis of the feet position, was contained below 0.03~m before being stabilized.

\subsection{Experimental results}

The same experiment was run on the physical iCub robot (with the difference that the robot lifted the left foot first rather than the right). The robot being new at the time the experiments were conducted, it needed further calibration and validation before achieving optimal results. The controller nonetheless proved to be effective for achieving one stride. Results achieved on the robot show that all tasks~\eqref{eq:tasks} were taken into account by the controller. 

\begin{figure}[!t]
    \centering
    
    \begin{subfigure}[b]{0.48\textwidth}
    \includegraphics[width=0.29\linewidth]{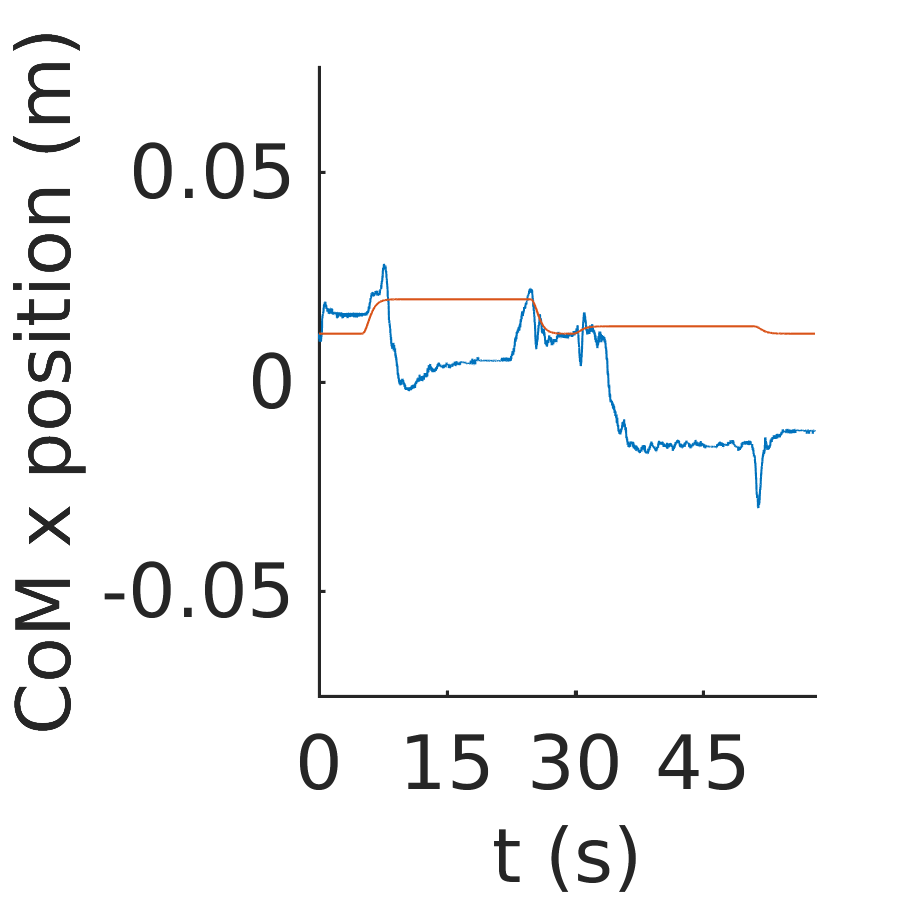}
    \includegraphics[width=0.29\linewidth]{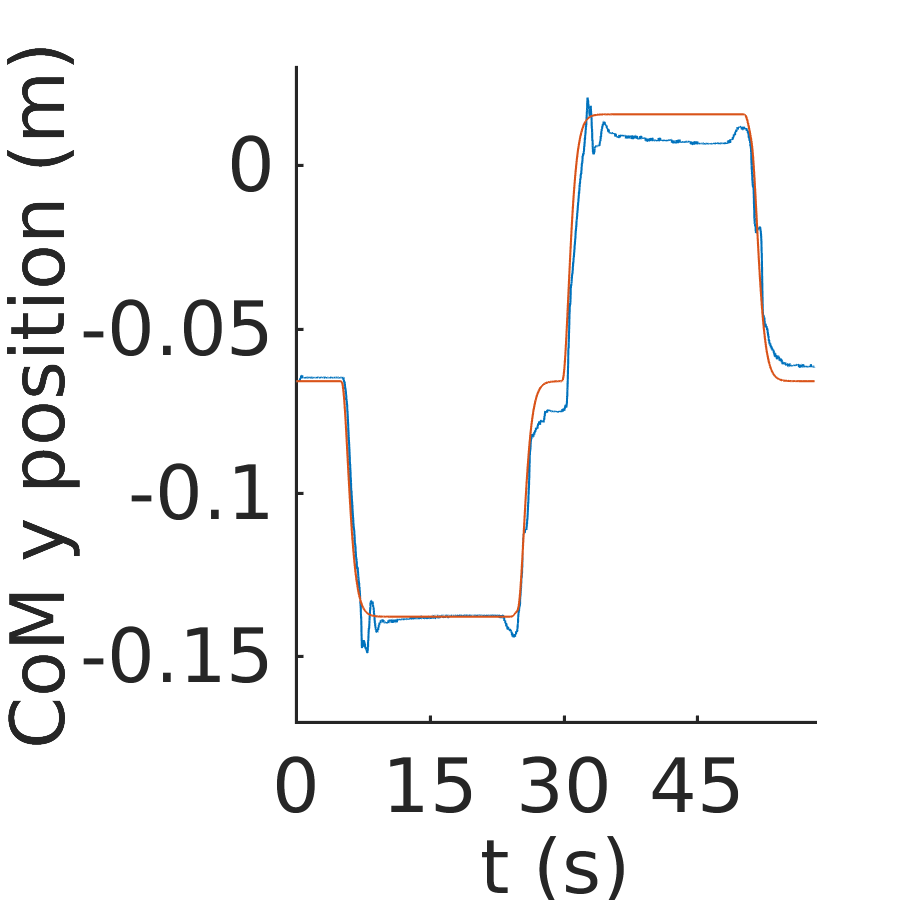}
    \includegraphics[width=0.29\linewidth]{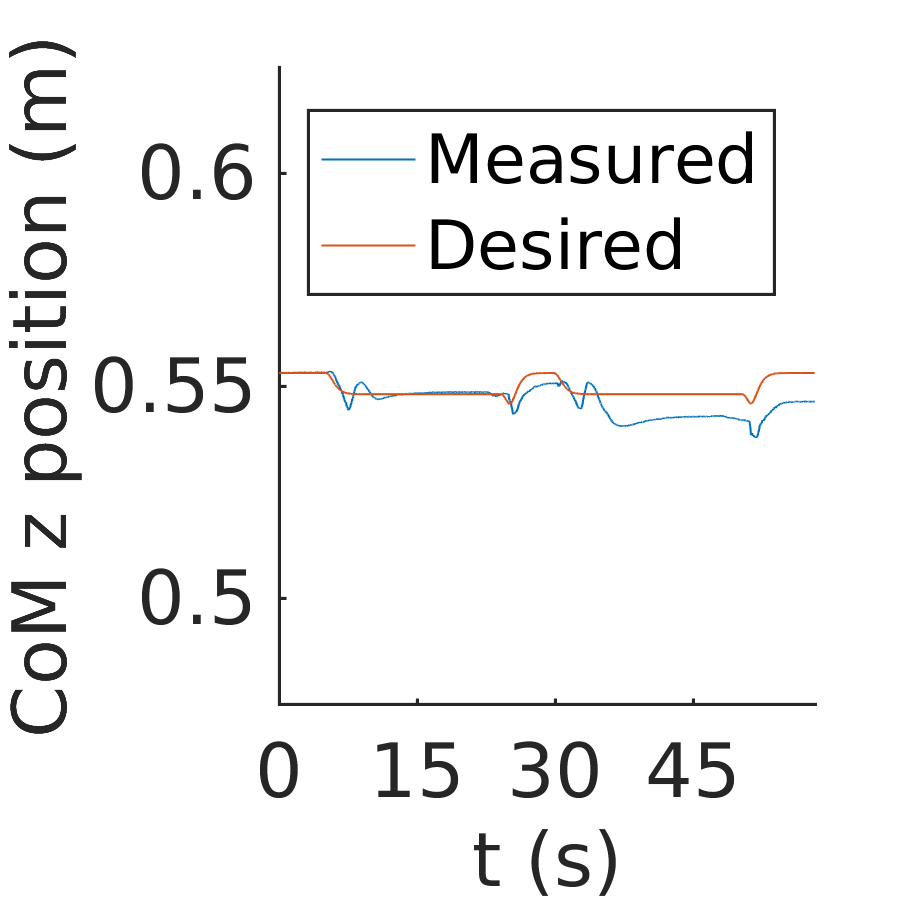}
    \caption{Center of mass (CoM) position}
    
    \includegraphics[width=0.29\linewidth]{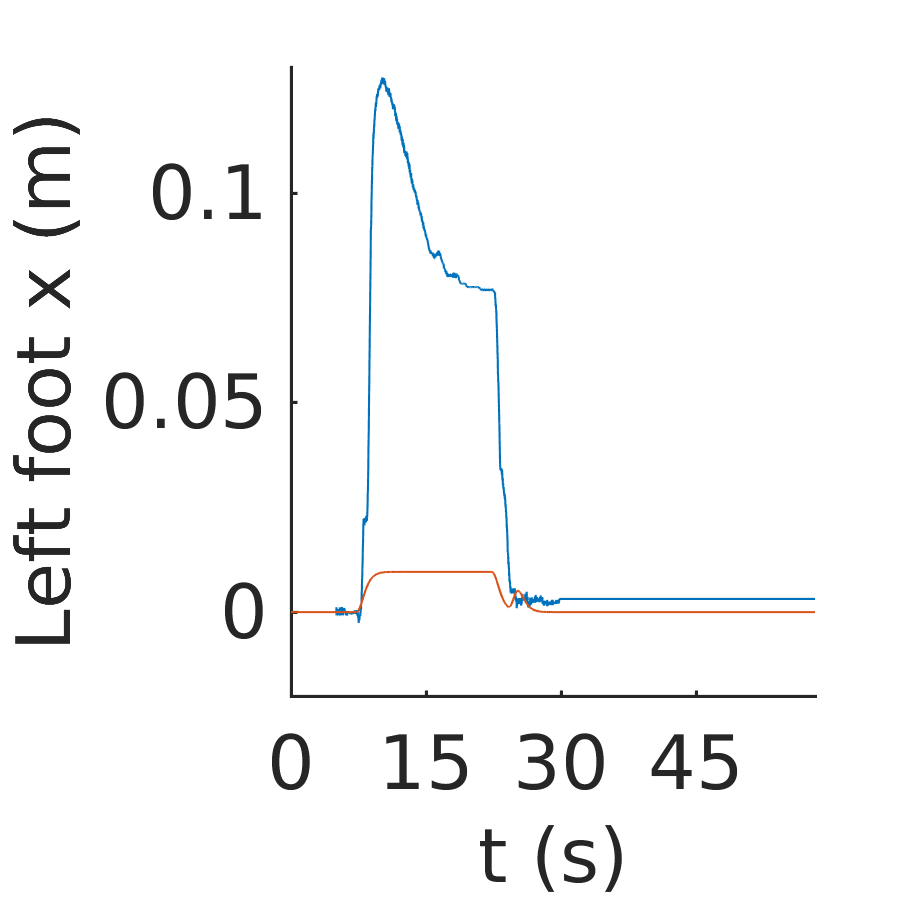}
    \includegraphics[width=0.29\linewidth]{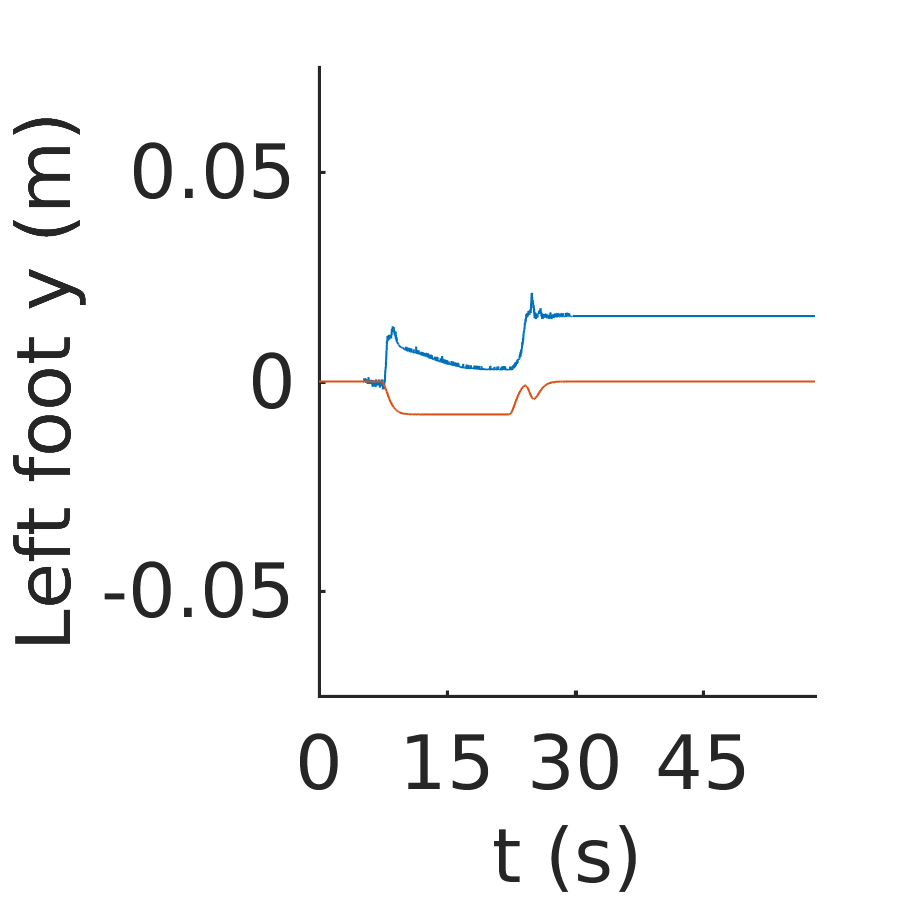}
    \includegraphics[width=0.29\linewidth]{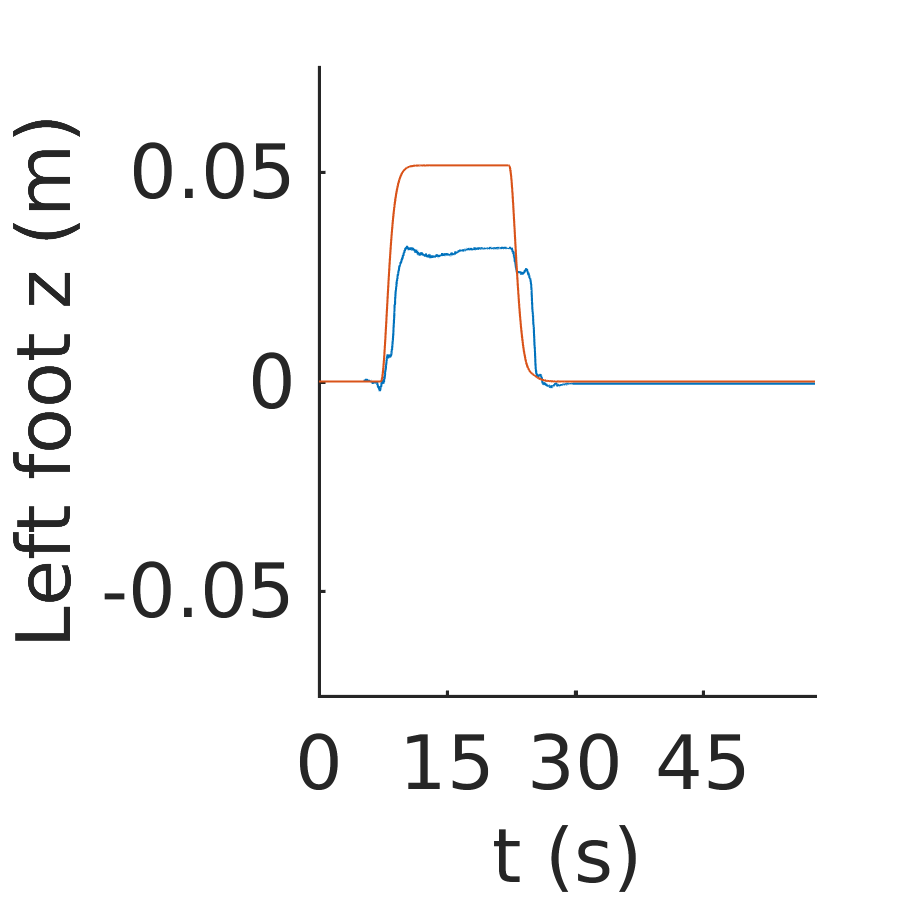}
    \caption{Left foot position}
    
    \includegraphics[width=0.29\linewidth]{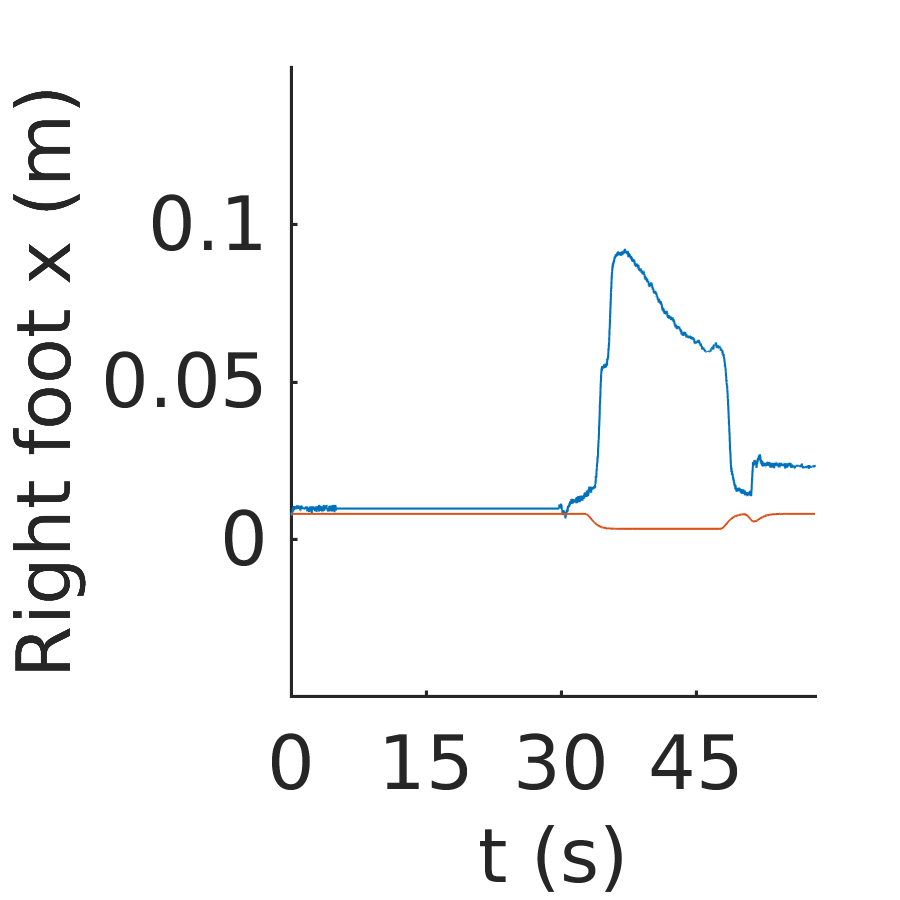}
    \includegraphics[width=0.29\linewidth]{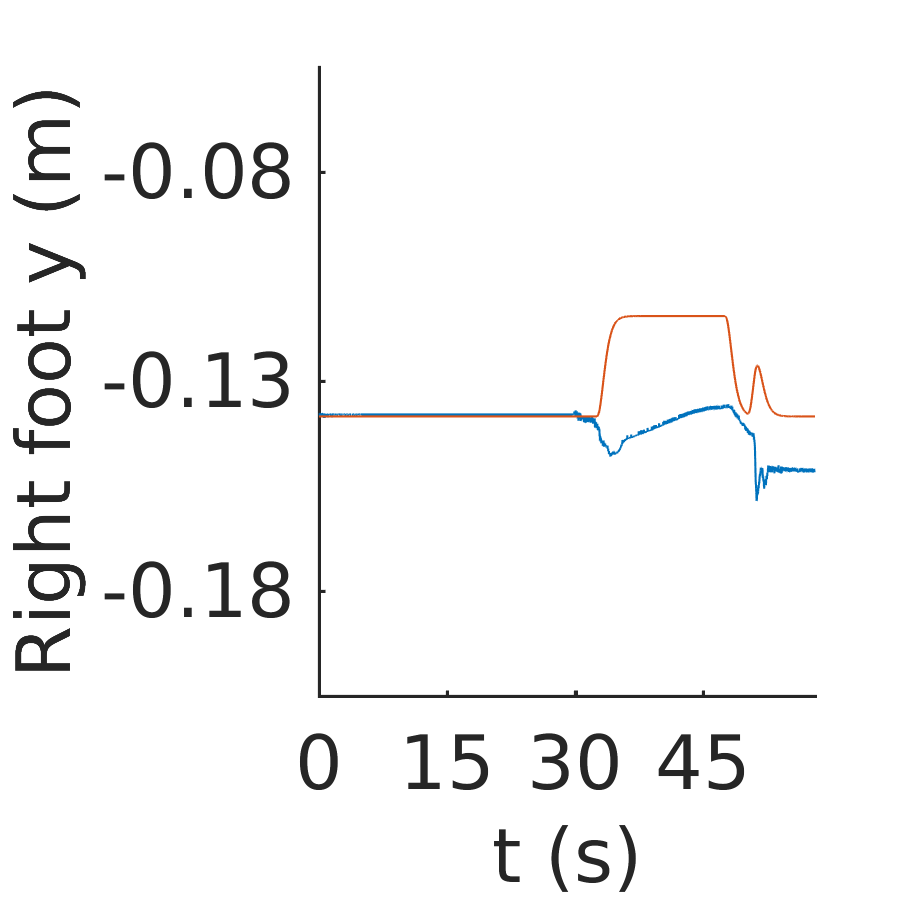}
    \includegraphics[width=0.29\linewidth]{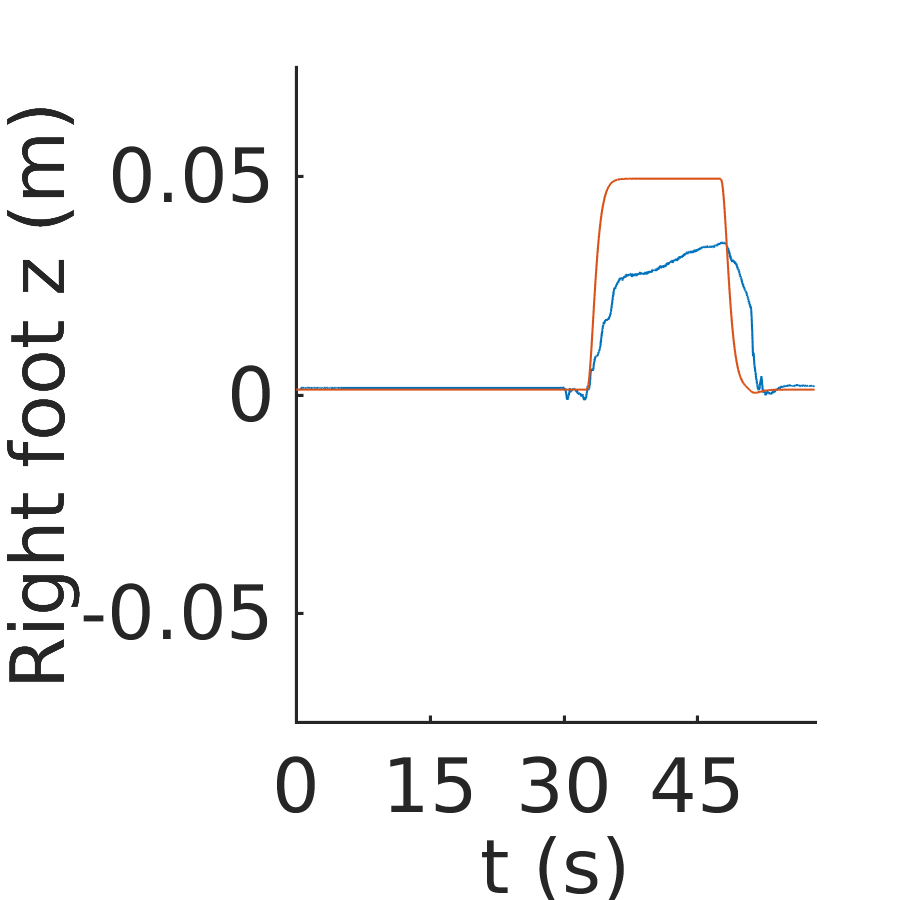}
    \caption{Right foot position}
    \end{subfigure}

    \caption{Evolution of position tasks for 1 stride performed with the robot. Position values are given with respect to a world frame of which the $x$, $y$ and $z$ axes correspond respectively to the sagittal, frontal and vertical axes. Achieved trajectories are shown in blue, while the desired ones are shown in red.}
    \label{fig:task_evolution_robot}
\end{figure}

The trajectories obtained for the center of mass and feet are shown in~\figref{fig:task_evolution_robot}. The error on the center of mass was generally kept below 0.02~m in each direction at all times, with the error on the $x$ axis being the largest. Regarding the feet positions, as in the simulation, they had a lower priority than the center of mass position. It was observed that at the beginning of the foot lifting movement, the foot was foot was first moved of about 0.1 m forward (due to the hip bending faster than the knee) before being brought back by the bending knee. At foot touchdown, each foot was brought back to its initial position with an error below 0.02 m. 

\begin{figure}[!t]
    \centering
    \begin{subfigure}[b]{0.32\linewidth}
        \includegraphics[width=\linewidth]{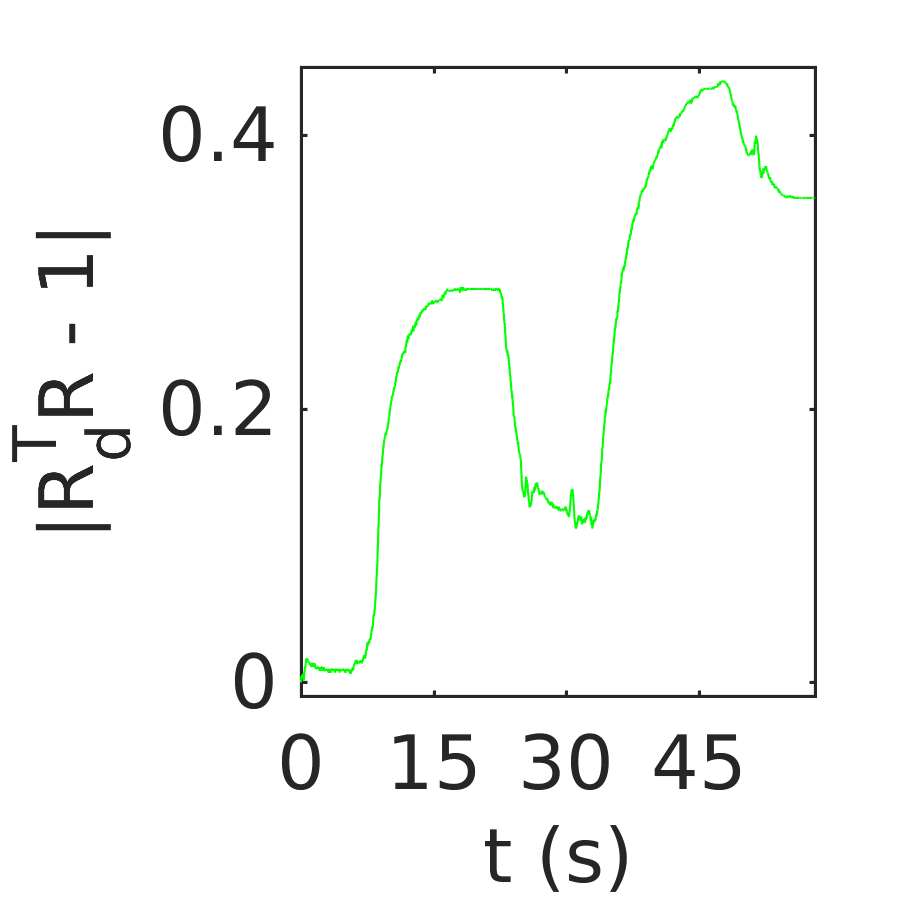}
        \caption{Root link}
    \end{subfigure}~
    \begin{subfigure}[b]{0.32\linewidth}   
        \includegraphics[width=\linewidth]{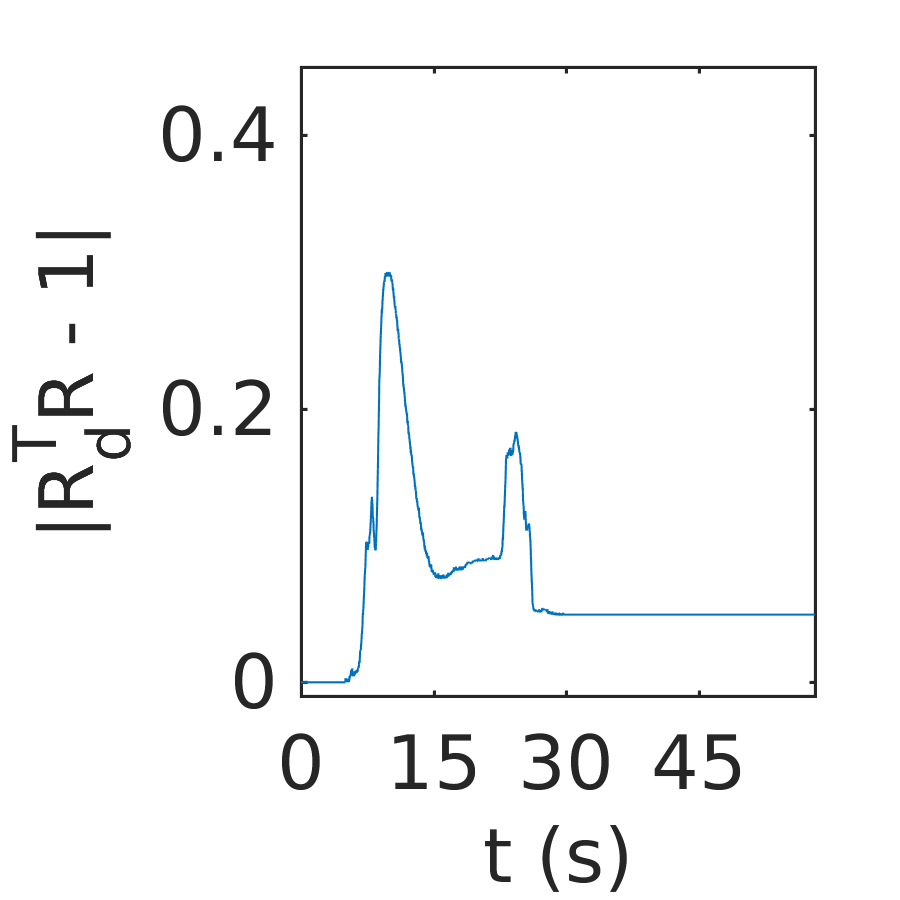}
        \caption{Left foot}
    \end{subfigure}~
     \begin{subfigure}[b]{0.32\linewidth}   
        \includegraphics[width=\linewidth]{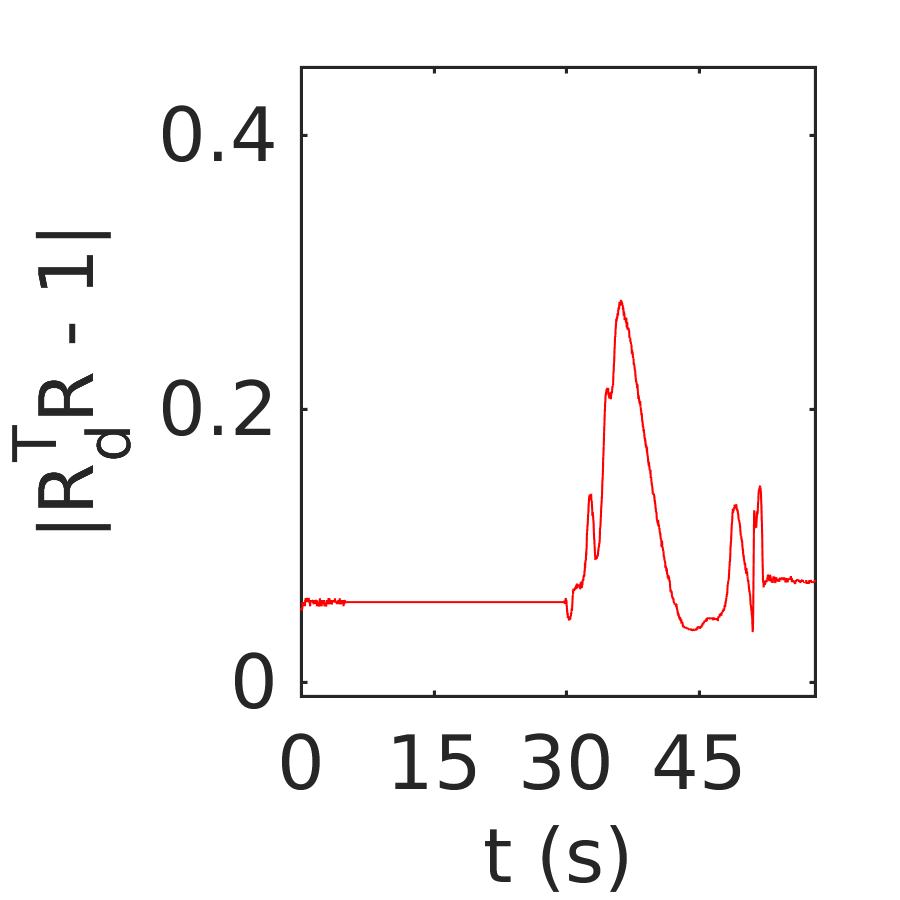}
        \caption{Right foot}
    \end{subfigure} 
    \caption{Evolution of orientation task errors for 1 stride performed with the robot. Achieved trajectories are shown in blue, while the desired ones are shown in red.}
    \label{fig:orientation_error_robot}
\end{figure}

The error obtained on the orientation tasks is shown in \figref{fig:orientation_error_robot}, by representing the error with $\left|R R_d^{\top} -1 \right|$. The obtained graphs show that the orientation error of the lifted feet is stabilized. The root orientation error is also stabilized, although it seems to increase for the second footstep; there may be a correlation with the similar increase on the center of mass position error.

\begin{figure}
    \centering
    \begin{subfigure}[b]{0.32\textwidth}
        \includegraphics[width=0.49\linewidth]{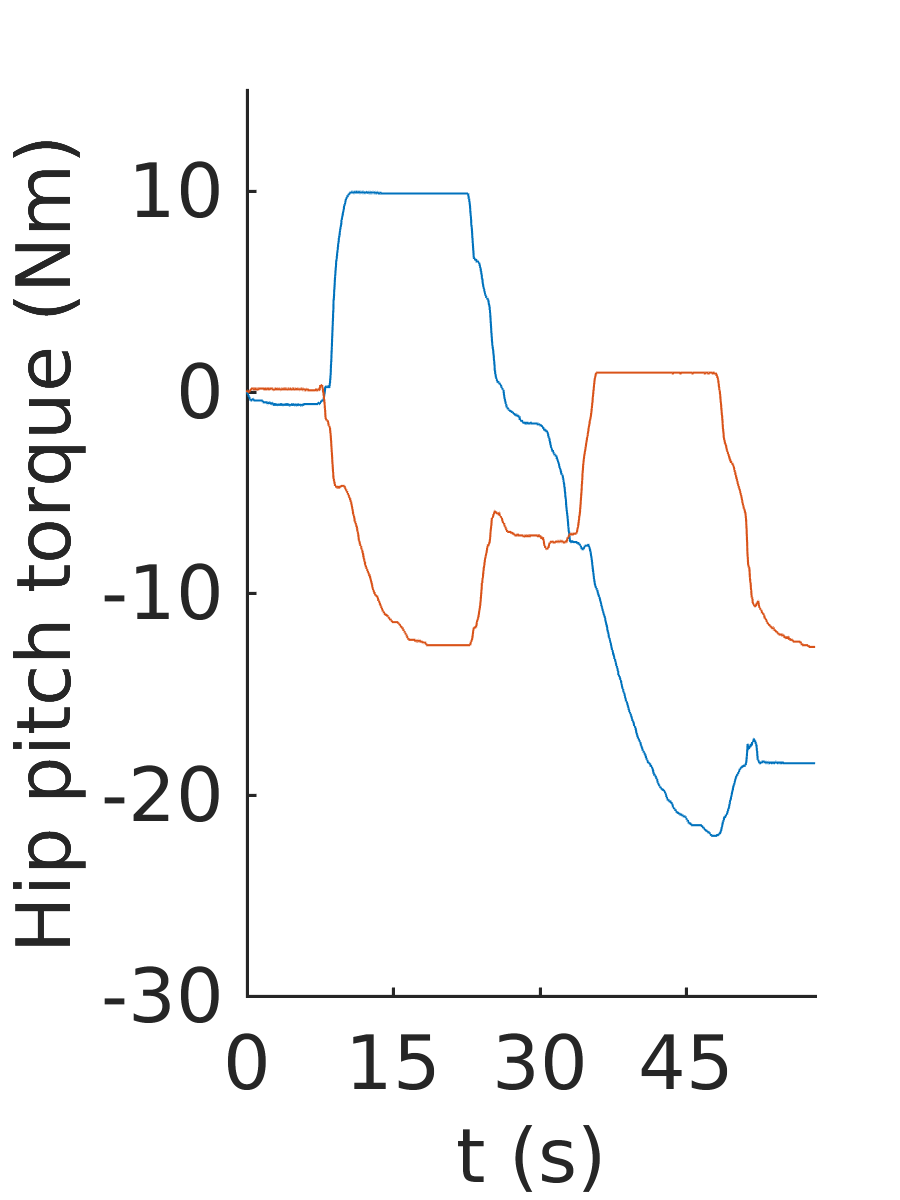}
        \includegraphics[width=0.49\linewidth]{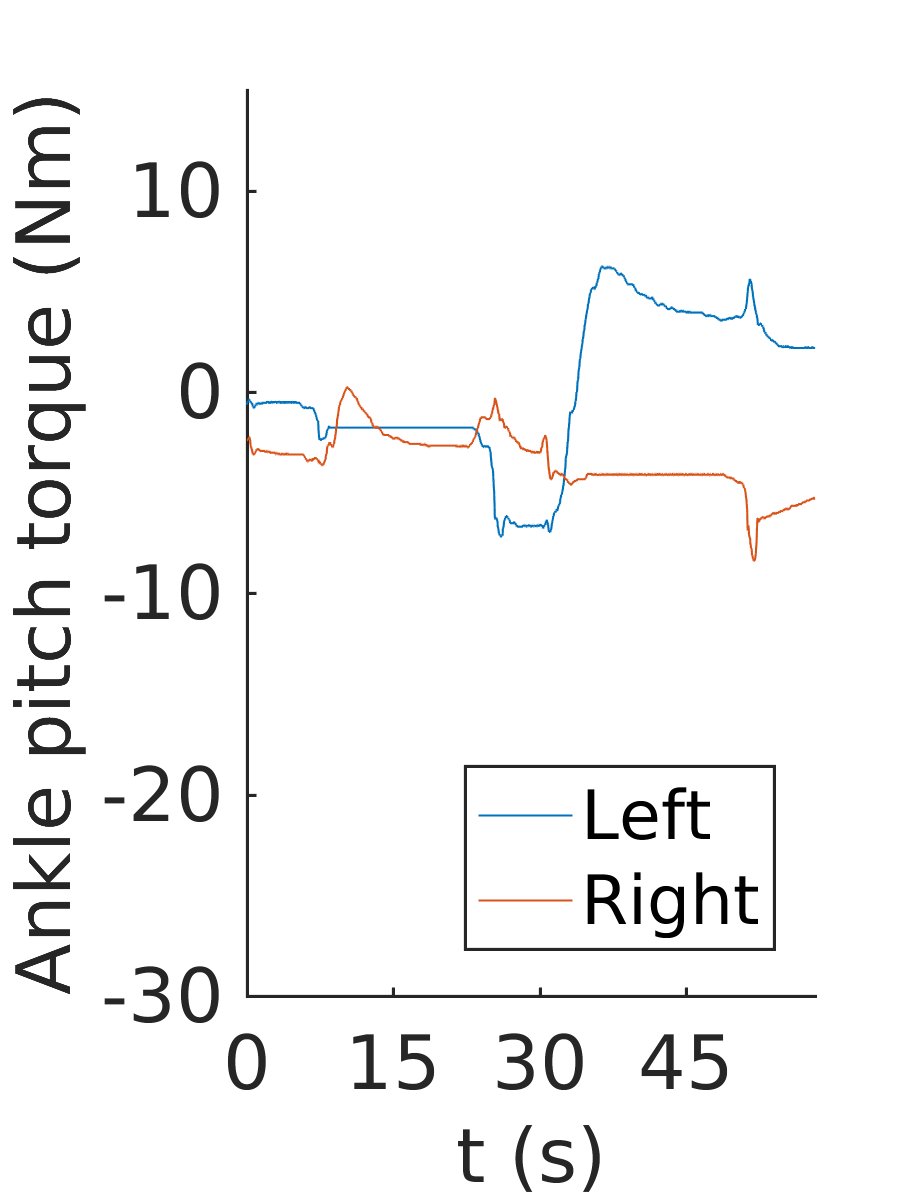}
         \caption{Torques about the hip and ankle pitch axis}
         
        \includegraphics[width=0.49\linewidth]{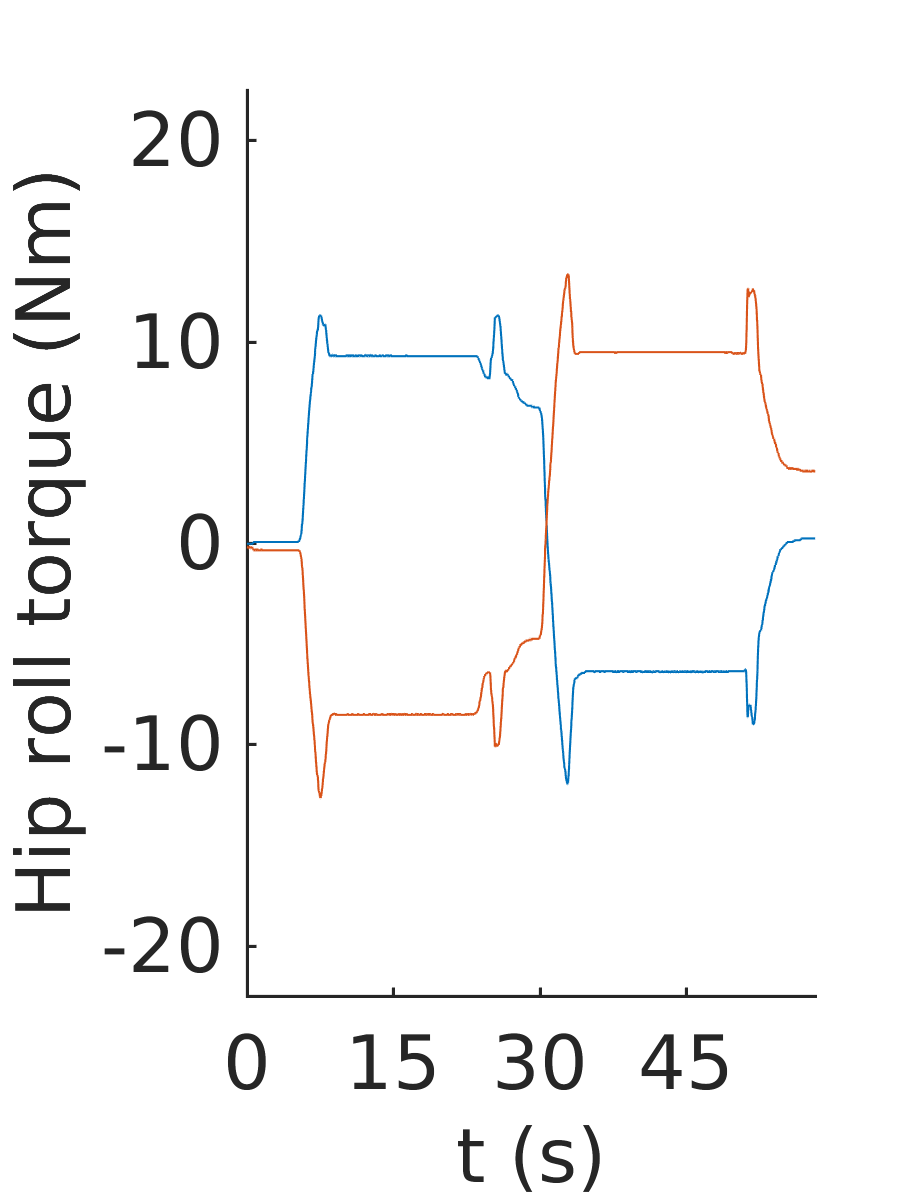}
        \includegraphics[width=0.49\linewidth]{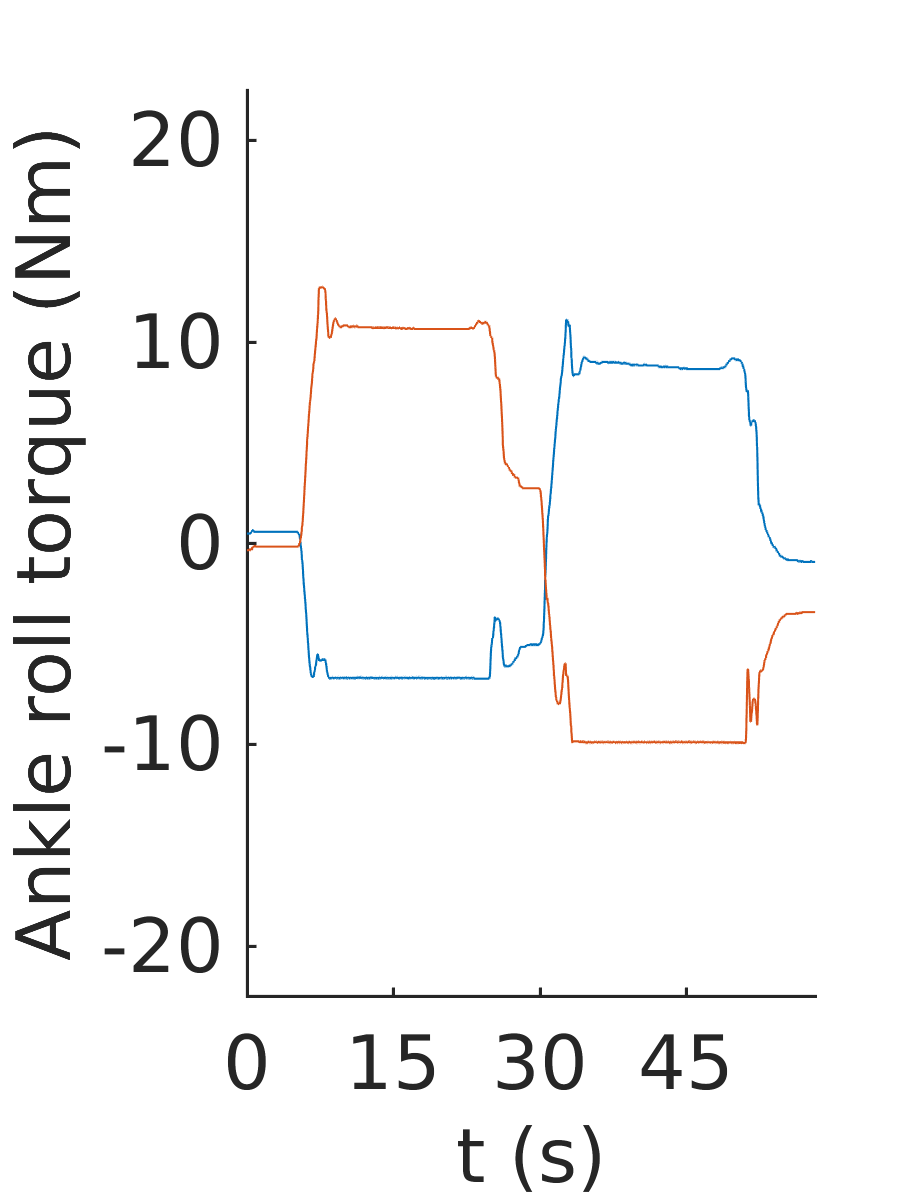}
        \caption{Torques about the hip and ankle roll axis}
    \end{subfigure}
    
    \caption{Evolution of torques on the robot for 1 stride, on the left and right hip and ankle joints. Torques on the pitch and roll axes of these joints were critical for keeping balance in the performed experiment.}
    \label{fig:torques_evolution_robot}
\end{figure}

Control torques obtained from the quadratic programming solver were directly applied to the robot. They are shown in~\figref{fig:torques_evolution_robot} for the joints which were most critical for balancing: the hips and ankles. The graphs show that torques were contained within a feasible range and relatively smooth, apart from the peaks which could be observed at foot liftoff and touchdown. 

\begin{figure}
    \centering
    \begin{subfigure}[b]{0.32\linewidth}
        \includegraphics[width=\linewidth]{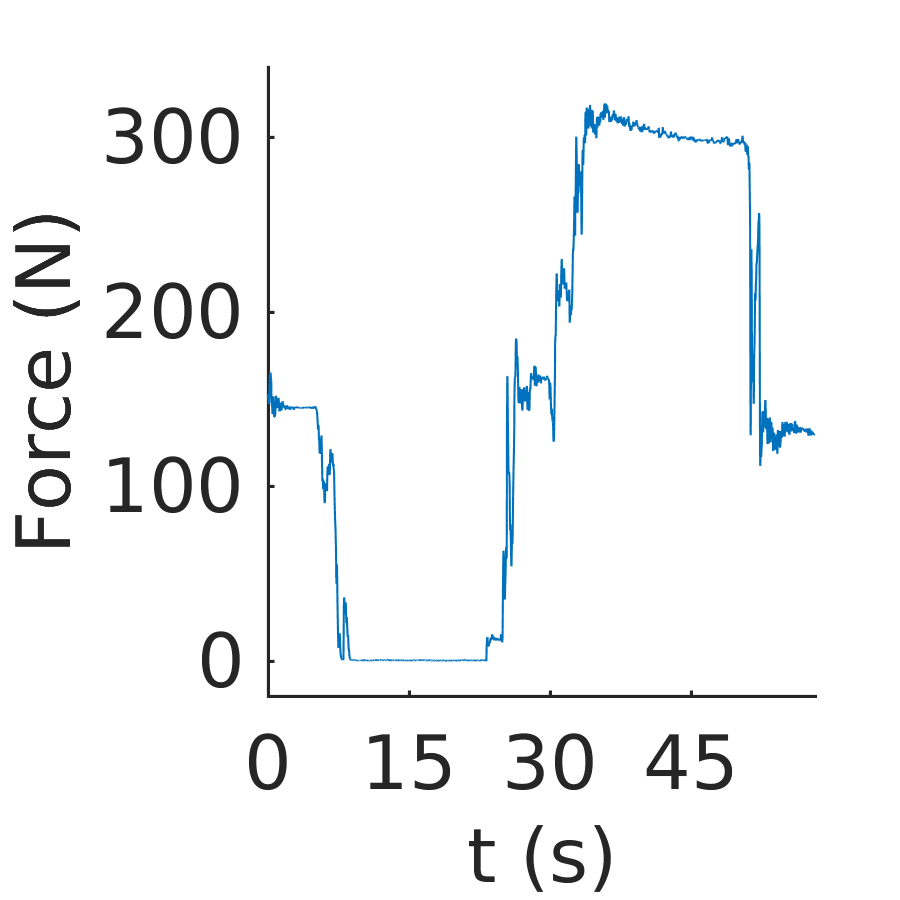}
        \caption{Left foot}
    \end{subfigure}~
    \begin{subfigure}[b]{0.32\linewidth}   
        \includegraphics[width=\linewidth]{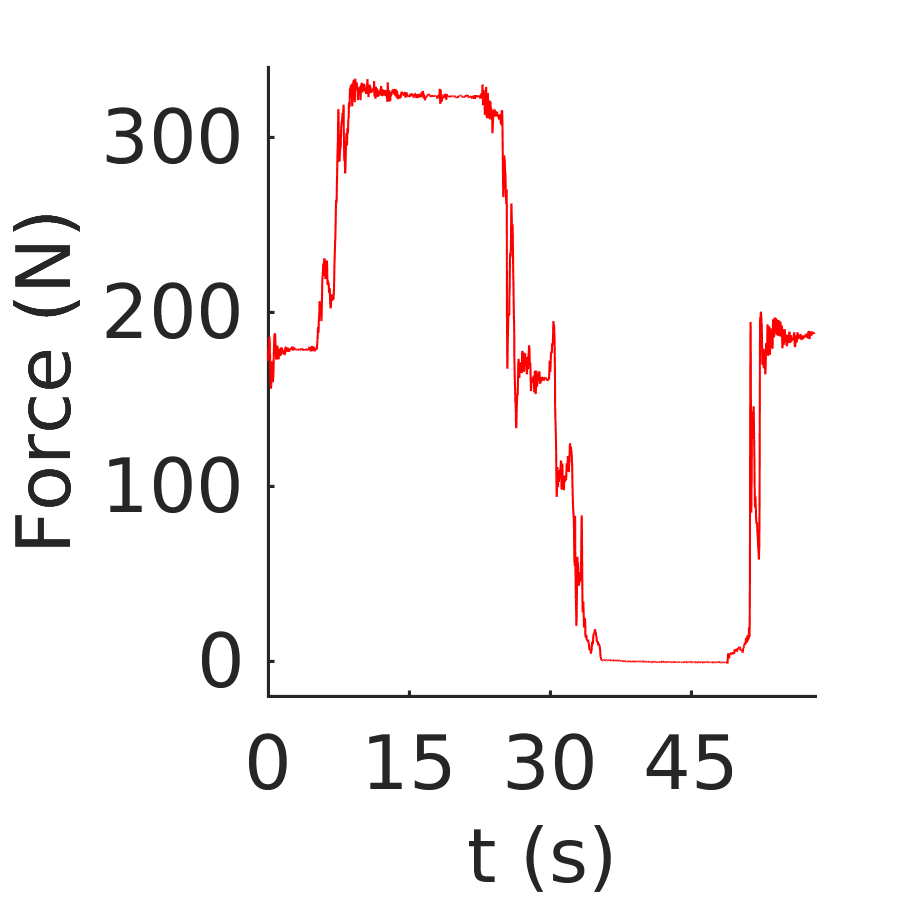}
        \caption{Right foot}
    \end{subfigure} 
    \caption{Evolution of the vertical contact forces applied at the robot feet for 1 stride.}    
    \label{fig:contact_wrench_robot}
\end{figure}

The vertical component of the resulting contact forces at the feet are shown in~\figref{fig:contact_wrench_robot}. At the moment of contact switching, although the magnitude of the forces varied rather rapidly, the forces do not show discontinuities which could not be handled by the robot.


Further experiments (including calibration, as well as parameter and gain tuning) are currently being conducted on the robot, in order to achieve a behavior closer to the one obtained in simulation.

\section{CONCLUSIONS} \label{sec:conclusions}


A control framework for whole-body torque control was proposed for balancing tasks. This algorithm was developed in order to enable walking, while allowing for physical interaction with the environment without loss of balance. It was implemented on the iCub humanoid robot to display the performance achieved with this method. Results obtained in simulation and on the robot show that all tasks were taken into account by the controller.

Through this work, it was shown that it is possible to achieve balancing and contact switching with a torque-controlled free-floating robot, without the use of centroidal momentum terms.

 Further experiments are being conducted, in order to achieve even better results on the robot.
 Although it was not emphasized in the paper, we found that the use of the low-level postural task helped enforce repeatability of the behavior of the robot. However, as it is formulated, the proposed controller assumes that input task and postural trajectories are feasible and coherent. For that reason, as future work, a trajectory planner should be designed in order to ensure such conditions. Following this, the controller shall be fit for dynamic walking trials. 







\section*{ACKNOWLEDGMENT}
Marie Charbonneau thanks Nuno Guedelha, Francisco Javier Andrade Chavez and Joan Kangro for alternately lending a hand when carrying out experiments on the robot.

 \newcommand{\noop}[1]{}

\end{document}